\documentclass[letterpaper]{article} % DO NOT CHANGE THIS
\usepackage{aaai2026}  % DO NOT CHANGE THIS
\usepackage{times}  % DO NOT CHANGE THIS
\usepackage{helvet}  % DO NOT CHANGE THIS
\usepackage{courier}  % DO NOT CHANGE THIS
\usepackage[hyphens]{url}  % DO NOT CHANGE THIS
\usepackage{graphicx} % DO NOT CHANGE THIS
\usepackage{natbib}  % DO NOT CHANGE THIS AND DO NOT ADD ANY OPTIONS TO IT
\usepackage{caption} % DO NOT CHANGE THIS AND DO NOT ADD ANY OPTIONS TO IT
\usepackage{algorithm}
\usepackage{algorithmic}

\usepackage{newfloat}
\usepackage{listings}
\DeclareCaptionStyle{ruled}{labelfont=normalfont,labelsep=colon,strut=off} % DO NOT CHANGE THIS
\floatstyle{ruled}
\newfloat{listing}{tb}{lst}{}
\floatname{listing}{Listing}
\usepackage{subfigure}
\usepackage{multirow}
\usepackage{makecell}
\usepackage{booktabs}
\usepackage{amsmath}
\usepackage{bm}
\usepackage{etoolbox}
\usepackage{amssymb}
\usepackage{pifont}

\newcommand{\greencheck}{{\ding{52}}}
\newcommand{\redcross}{{\ding{55}}}

\title{ConInstruct: Evaluating Large Language Models on \\Conflict Detection and Resolution in Instructions}
\author{
    Xingwei He\textsuperscript{\rm 1},
    Qianru Zhang\textsuperscript{\rm 1}\thanks{Corresponding authors.},
    Pengfei Chen\textsuperscript{\rm 2}\footnotemark[1],\\
    Guanhua Chen\textsuperscript{\rm 3},
    Linlin Yu\textsuperscript{\rm 4},
    Yuan Yuan\textsuperscript{\rm 5,6,7},
    Siu-Ming Yiu\textsuperscript{\rm 1}
}
\affiliations{
    \textsuperscript{\rm 1}The University of Hong Kong,
    \textsuperscript{\rm 2}Xidian University,
    \textsuperscript{\rm 3}Southern University of Science and Technology,
    \textsuperscript{\rm 4}Augusta University
    \textsuperscript{\rm 5}Beihang University,
    \textsuperscript{\rm 6}Qingdao Research Institute, Beihang University, 
    \textsuperscript{\rm 7}Hangzhou Innovation Institute, Beihang University\\
    hexingwei15@gmail.com,
    qrzhang@cs.hku.hk,
    chenpengfei@xidian.edu.cn
}

\usepackage{bibentry}
\begin{document}

\maketitle

\begin{abstract}
Instruction-following is a critical capability of Large Language Models (LLMs). While existing works primarily focus on assessing how well LLMs adhere to user instructions, they often overlook scenarios where instructions contain conflicting constraints—a common occurrence in complex prompts. The behavior of LLMs under such conditions remains under-explored. 
To bridge this gap, we introduce ConInstruct, a benchmark specifically designed to assess LLMs' ability to detect and resolve conflicts within user instructions. Using this dataset, we evaluate LLMs' conflict detection performance and analyze their conflict resolution behavior. 
Our experiments reveal two key findings: (1) 
Most proprietary LLMs exhibit strong conflict detection capabilities, whereas among open-source models, only DeepSeek-R1 demonstrates similarly strong performance. DeepSeek-R1 and Claude-4.5-Sonnet achieve the highest average F1-scores at 91.5\% and 87.3\%, respectively, ranking first and second overall. 
(2) Despite their strong conflict detection abilities, LLMs rarely explicitly notify users about the conflicts or request clarification when faced with conflicting constraints. 
These results underscore a critical shortcoming in current LLMs and highlight an important area for future improvement when designing instruction-following LLMs.
Our code and dataset are available at: \url{https://github.com/NLPCode/ConInstruct}.
% Our dataset spans six diverse tasks, with each instruction containing six types of constraints and seven to nine types of conflicts. 

% tend to generate responses without explicitly informing users about the presence of conflicting constraints. 
% (3) As the number of conflicts in instructions increases, strong LLMs exhibit two notable behavioral patterns: (a) explicitly notifying users about the conflicts and requesting clarification, or (b) acknowledging the conflicts, autonomously resolving them, and responding to the resolved instructions.
\end{abstract}

%     \link{Code and Dataset}{github.com/NLPCode/ConInstruct}
%     % \link{Datasets}{https://aaai.org/example/datasets}
%     % \link{Extended version}{https://aaai.org/example/extended-version}

\section{Introduction}
Large Language Models (LLMs) \cite{Achiam2023GPT4TR, touvron2023llama, PaLM} 
have witnessed significant advancements in recent years, 
demonstrating remarkable capabilities in reasoning \cite{wei2022chain, wang2022self}, and time-series forecasting \cite{jia2024gpt4mts, Zhang2024ASO, zhang2025autohformer}. 
A fundamental ability of LLMs is to follow instructions—generating responses that align with user-provided instructions. 
Instruction-following \cite{ouyangtraining} has emerged as a key research focus, playing a critical role in enhancing the interpretability, controllability, and trustworthiness of LLMs. 

Existing instruction-following works primarily focus on evaluating to what extent LLMs' outputs align with user instructions using rule-based and model-based evaluation methods. 
For rule-based evaluation, \citet{zhou2023instruction} proposed IFEval, a benchmark comprising verifiable instructions (e.g., ``\textit{Include the keyword `useful' in your response}''), where a rule-based program can verify whether a model's output meets the given instructions. Meanwhile, recent studies suggest that LLMs can rival human annotators \cite{he-etal-2024-annollm} and serve as reliable evaluators \cite{NEURIPS2023_91f18a12}. Building on these findings, model-based evaluation \cite{Chen_Xu_Wang_Liu_Mao_2024, qin-etal-2024-infobench} leverages strong LLMs to automatically assess whether LLMs' outputs adhere to user instructions.  The latest research integrates rule-based and model-based evaluation approaches \cite{jiang-etal-2024-followbench, zhang2024cfbench, wen2024benchmarking}. On the other hand, concurrent works \cite{wallace2024instruction, zhang-etal-2025-iheval, geng2025control} evaluate whether LLMs can follow an instruction hierarchy, where high-level instructions (e.g., system instructions) take precedence over low-level ones (e.g., user instructions). 

\begin{figure}
    \centering
    \includegraphics[width=0.47\textwidth]{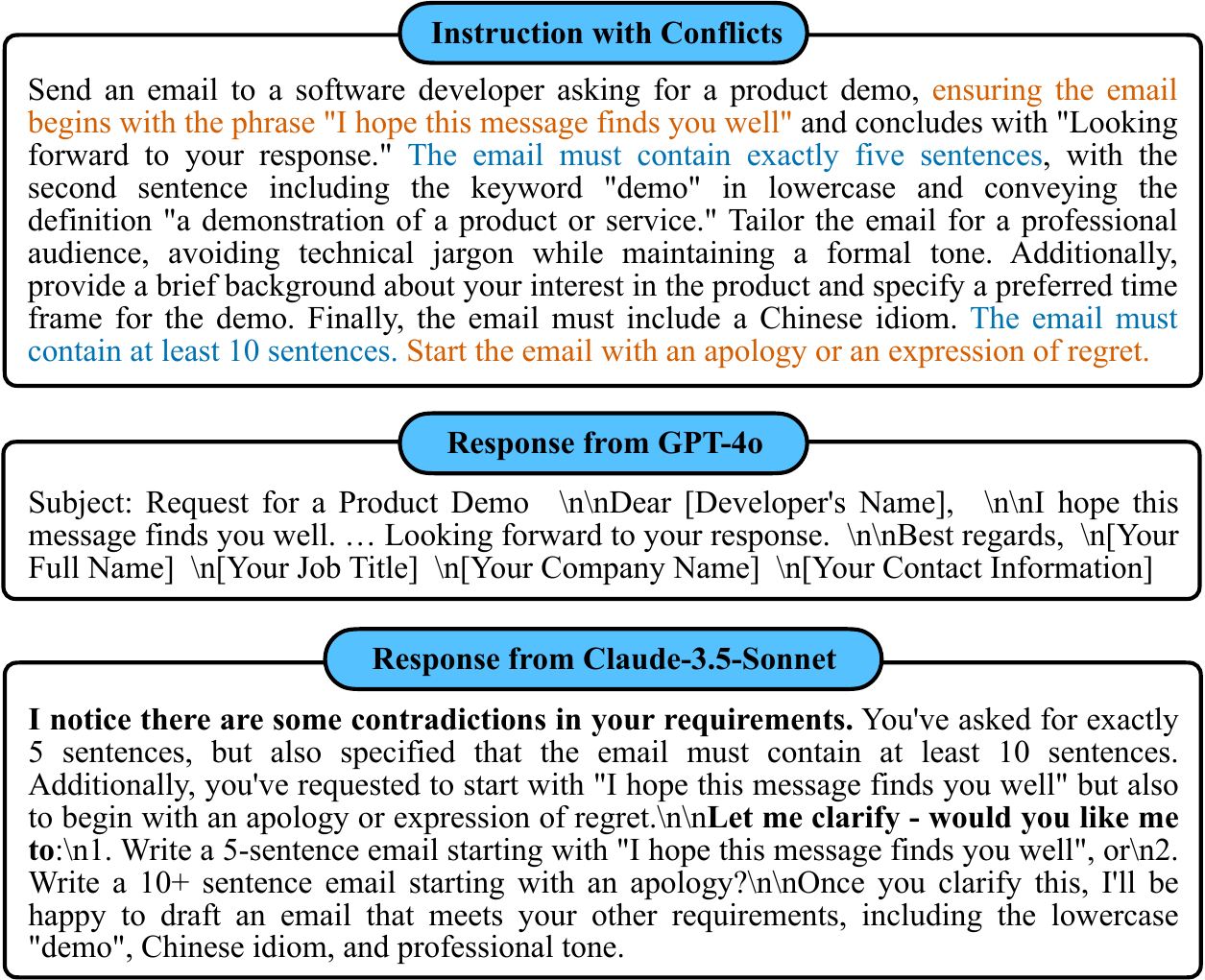} 
    \caption{ 
    An instruction with conflicts from ConInstruct, where text in green and red indicate conflicts between phrase constraints and length constraints, respectively. The lower part of the figure presents two responses from GPT-4o and Claude-3.5-Sonnet for the instruction.
    }
    \label{fig.example}
\end{figure}

Prior works assume that all constraints in the user instructions are coherent and non-conflicting. In practice, when users provide long or complex instructions, they may unintentionally introduce conflicting constraints—requirements that cannot be simultaneously satisfied by LLMs. 
Figure \ref{fig.example} illustrates an instruction containing two conflicts: one between phrase constraints and another involving length constraints. The presence of such conflicts poses a unique challenge for LLMs. 
If an LLM generates a response without notifying the user of these conflicts (as seen in GPT-4o's response in Figure \ref{fig.example}), the user may not realize that their instruction contains conflicts and the model's output fails to fully satisfy the instruction.
In such cases, a preferable conflict resolution behavior is to explicitly inform the user about the conflicts and request clarification before proceeding (as shown in Claude-3.5-Sonnet's response in Figure \ref{fig.example}). 
Despite the growing interest in instruction-following, no prior work has systematically evaluated LLMs' performance when faced with user instructions with conflicts.

\begin{figure*}
    % \vspace{-20mm}
    \centering
    \includegraphics[width=0.98\textwidth]{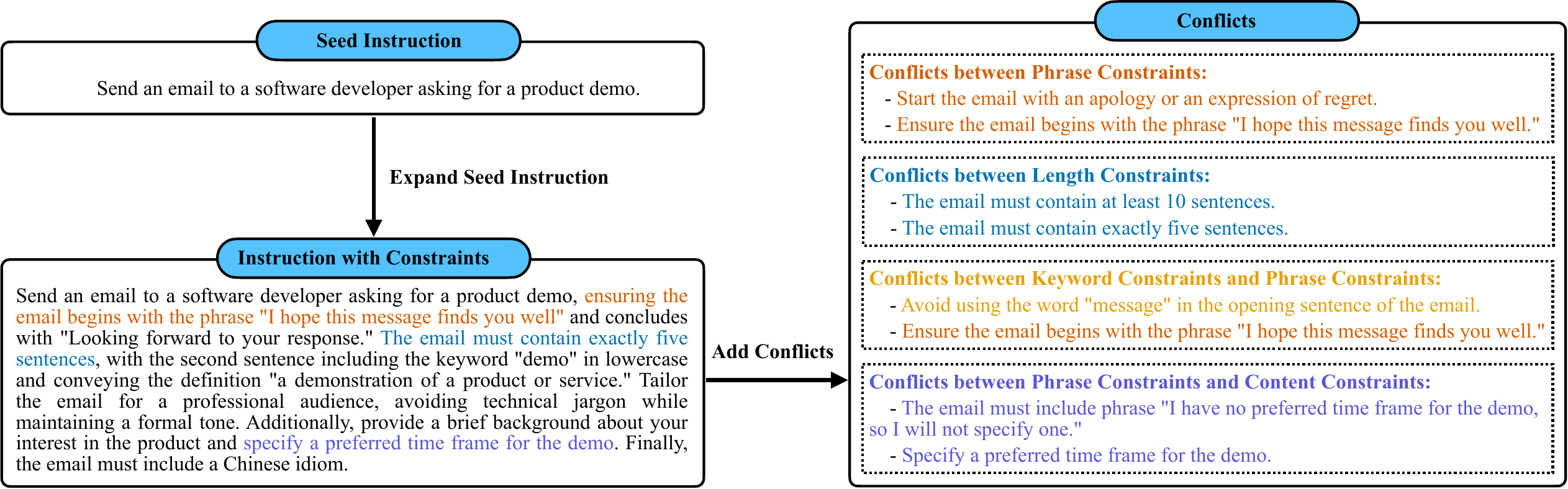} 
    % \vspace{-3mm}
    \caption{ 
    The construction process of the ConInstruct Benchmark: We start with a seed instruction, then add constraints to it. Finally, we introduce conflicts into the expanded instructions. Due to space limits, we show only four conflicts. In each conflict, the first constraint is newly added, while the second comes from the original instruction. 
    % These two constraints conflict with each other.
    }
    % \vspace{-4mm}
    \label{fig.data_construction}
\end{figure*}

To bridge this gap, we introduce \textbf{ConInstruct}, 
% \footnote{Our code and dataset are available at: https://github.com/NLPCode/ConInstruct.}, 
a novel dataset designed to evaluate LLMs on \textbf{Con}flicting \textbf{Instruct}ions that contain diverse constraints. 
Specifically, our dataset covers six distinct tasks, with each instruction incorporating six types of constraints: content, keyword, phrase, length, format, and style constraints. Furthermore, we design 7-9 different types of conflicts per instruction, including both intra-constraint conflicts (e.g., conflicts between phrase constraints) and inter-constraint conflicts (e.g., conflicts between keyword and phrase constraints) (please refer to conflicts in Figure \ref{fig.data_construction}). 
Using this dataset, we systematically analyze LLMs' performance in \textbf{conflict detection} and examine their behaviors in \textbf{conflict resolution}. 

Conflict detection assesses how well LLMs can identify conflicts within a given instruction. 
To evaluate this, we introduce a new constraint into a conflict-free instruction, ensuring it conflicts with an already present constraint. 
We then ask LLMs to determine whether the instruction contains conflicting constraints. 
Our results show that proprietary LLMs exhibit strong conflict detection capabilities, with Claude-4.5-Sonnet achieving the second-highest average F1-score at 87.3\%.
% These findings suggest that advanced LLMs can effectively detect conflicts in user instructions. 
Notably, as the number of conflicts in an instruction increases, LLMs exhibit improved conflict detection ability, aligning with our intuitions.

Conflict resolution, on the other hand, investigates how LLMs behave when faced with instructions containing conflicts. 
While LLMs perform well in conflict detection, our findings indicate that they often generate responses without explicitly informing the user about conflicts. 
For example, when an instruction contains 1–2 conflicts, GPT-4o will directly generate a response in 97.5\% of cases, satisfying only a subset of the constraints but failing to notify the user of the conflicts. 
Even the best-performing model, Claude-4.5-Sonnet, explicitly alerts users to conflicts in only 45\% of cases—either by (1) requesting further clarification (36\%) or (2) resolving the conflicts autonomously and responding to the resolved instruction (9\%).  
Moreover, as the number of conflicts in an instruction increases, strong LLMs (Claude models and GPT-4o) become more likely to acknowledge the existence of conflicts in their responses.

Our contributions can be summarized as follows: 
(1) We introduce ConInstruct, a novel dataset designed to evaluate LLM performance in handling user instructions with conflicts. 
(2) We conduct an in-depth study on conflict detection, demonstrating that proprietary LLMs exhibit strong detection capabilities. 
(3) We analyze the conflict resolution behaviors exhibited by LLMs when encountering conflicting instructions. Our findings reveal that while proprietary LLMs exhibit strong conflict detection capabilities, they often fail to convey conflicts explicitly in their responses, highlighting an important area for future improvement in instruction-following LLMs. 
% (4) Our findings provide valuable insights into the limitations of current LLMs in handling conflicting instructions, paving the way for future research on enhancing instruction-following reliability.

\section{ConInstruct Benchmark}

\subsection{Dataset Construction}
As shown in Figure \ref{fig.data_construction}, the construction of ConInstruct consists of three steps: preparing seed instructions, expanding them with constraints, and introducing conflicts into the expanded instructions. 
In the following, we will provide further details on each step.

\paragraph{Preparing Seed Instructions.}
We begin by manually curating 100 seed instructions, which serve as fundamental instructions without additional constraints. In designing these seed instructions, we prioritize task and domain diversity to ensure broad coverage across various scenarios. 
To be specific, ConInstruct comprises six common NLP tasks: \textit{email writing, plan generation, story generation, open-domain question answering (QA), review writing, and article writing}. These tasks span 35 scenario-specific domains, including \textit{travel, work, health, finance, technology, and history}. 
We present the task and domain distribution for ConInstruct in Figure 7 of Appendix B. 
Overall, the seed instructions provide a diverse set of tasks and scenarios.
% Overall, ConInstruct provides a diverse set of tasks and scenarios, enabling a more comprehensive evaluation of LLMs' performance when handling conflicting instructions.

\paragraph{Constraint Types.}
Following previous work on instruction-following \cite{jiang-etal-2024-followbench, he2024can}, we utilize six widely-used types of constraints to expand the seed instructions. 
\textbf{Content Constraints} require the output to include specific details related to the content, such as reasons, purposes, topics, or background information. 
\textbf{Keyword Constraints} enforce the inclusion of specific keywords in the output or specify constraints on their part of speech or meaning. 
\cite{he-yiu-2022-controllable}. 
\textbf{Phrase Constraints} mandate the presence of specific phrases or sentences in the output.  
\textbf{Length Constraints} impose restrictions on the length of the output, such as word count, sentence count, or paragraph count. 
\textbf{Format Constraints} specify the format of the output (e.g., JSON, Markdown) or its language format (e.g., requiring the output to be entirely in English). 
\textbf{Style Constraints} control aspects such as sentiment, readability, and overall tone of the output. 
Further details on these constraint types are provided in Appendix C.

\paragraph{Expanding Seed Instructions.} 
We leverage GPT-4o to inject constraints into seed instructions. To enhance constraint diversity, we require GPT-4o to incorporate all six types of constraint into each seed instruction. See Figure \ref{fig.data_construction} for an example of an expanded instruction. The prompt used for this expansion is detailed in Table 4 of Appendix E.

\paragraph{Conflict Types.} 
When designing conflicting constraints, we prioritized the feasibility of evaluating constraint satisfaction using LLMs or automated programs. To this end, we define nine types of conflicts based on six widely used constraints \cite{jiang-etal-2024-followbench, he2024can}, categorized into six \textbf{intra-constraint conflicts} and three \textbf{inter-constraint conflicts}.
% Before introducing conflicts into the expanded instructions, we define nine conflict types, categorized into six \textbf{intra-constraint conflicts} and three \textbf{inter-constraint conflicts}. 
Intra-constraint conflicts occur within the same constraint type, including conflicts within Content Constraints (\textbf{CC}), Keyword Constraints (\textbf{KK}), Phrase Constraints (\textbf{PP}), Length Constraints (\textbf{LL}), Format Constraints (\textbf{FF}), and Style Constraints (\textbf{SS}). 
Inter-constraint conflicts occur between different constraint types, including conflicts between Keyword and Phrase Constraints (\textbf{KP}), Phrase and Content Constraints (\textbf{PC}), and Phrase and Style Constraints (\textbf{PS}). 
Further details are provided in Appendix D.

\begin{table*}[!t]
    \footnotesize
    \centering
    % \resizebox{0.8\textwidth}{!}{%
    % \setlength{\tabcolsep}{1.5pt}% column space
    \begin{tabular}{ccccc|ccccccccc}
    % \toprule
    \hline
    \multicolumn{5}{c}{\textbf{Basic Statistics}}&  \multicolumn{9}{c}{\textbf{Conflict Distribution}}\\
    % \midrule
    \hline
    \textbf{Inst.} & \textbf{Word} & \textbf{Sent.} &  \textbf{CT}& \textbf{CFT}&\textbf{CC}&\textbf{KK}&\textbf{PP}&\textbf{LL}&\textbf{FF}  &\textbf{SS}&\textbf{KP}&\textbf{PC}&\textbf{KP}\\
    \hline
    % \midrule
     100& 138.9 & 6.4 &6 & 8.6 &100 & 100 & 100 & 100 & 100 & 100 & 100 & 94 & 70\\
    \hline
    % \bottomrule
    \end{tabular}
    % }
    % \vspace{-3mm}
    \caption{ConInstruct Statistics. `Inst.', `Word', `Sent.', `CT', and `CFT'  denote the number of expanded instructions, average words, sentences, constraint types, and conflict types per instruction. The right half of the table shows the number of conflicts for each conflict type.}
    % \vspace{-4mm}
    \label{table.data_statistics}
\end{table*}

\paragraph{Adding Conflicts.} We use GPT-4o to introduce conflicting constraints into the expanded instructions. To better control the number of conflicts in each instruction, we prompt the model to generate conflict pairs rather than directly injecting conflicting constraints into the instructions. Each conflict pair consists of two constraints: one extracted from the expanded instruction and another, newly constructed by GPT-4o, that directly contradicts the former. We instruct GPT-4o to generate one conflict pair for each of the nine predefined conflict types. 
Figure \ref{fig.data_construction} illustrates four conflict pairs corresponding to an expanded instruction. The prompt used to add conflicts is provided in Table 5 of Appendix E.

\subsection{Quality Control}
To ensure the data quality of ConInstruct, we use a two-step verification process for each instruction. 
In the first step, two annotators refine the expanded instructions and conflicts generated by GPT-4o. For expanded instructions, they assess the reasonableness and correctness of constraints, correcting any unreasonable or erroneous ones. They also check whether the expanded instructions include all six types of constraints and add any missing ones. 
For conflicts, annotators examine whether newly introduced constraints are indisputably in conflict with the constraints in expanded instructions. Any ambiguous conflicts are revised accordingly. For example, if the constraint in an expanded instruction states that ``The email should contain 150–200 words'', and a new constraint states that ``The email must be brief,'' the conflict is ambiguous because ``brief'' lacks a clearly defined word limit. Annotators also ensure that all types of conflicts are covered and construct any missing ones. 
In the second step, a third annotator\footnote{All annotators are college students and independent of our research.} reviews the revised instructions and conflicts, removing unreasonable constraints or conflicts.

\subsection{Dataset Statistics}\label{section.data_statistics}
Table \ref{table.data_statistics} presents the basic statistics of the expanded instructions in ConInstruct. Each instruction contains six types of constraints and an average of 8.6 conflict types. In the conflict detection and resolution experiments, we construct conflicting instructions by combining conflicts with expanded instructions. Specifically, we append the new constraints from the conflicts directly to the end of the expanded instructions. This approach allows us to generate a sufficient number of instructions with varying numbers of conflicts. For example, when the number of conflicts is set to one, we can construct a total of 864 conflicting instructions.

\section{Experiment Setup}
We will introduce the common experiment setup for conflict detection and conflict resolution.
\subsection{Preparing Instructions with Conflicts}
For each task, we first evaluate LLMs on instructions with a single conflict and then analyze their behaviors on instructions with multiple conflicts.

\paragraph{Instructions with a Single Conflict.} 
As introduced in Section \ref{section.data_statistics}, each expanded instruction contains \( n \) different types of conflicts (\( 7 \leq n \leq 9 \)). 
For each instruction \( I_i \in \mathcal{I}_0 \) (\( \mathcal{I}_0 \) denotes the set of conflict-free expanded instructions from ConInstruct) and its corresponding conflicts \( \{c_1, c_2, \dots, c_n\} \), we append each conflict to \( I_i \), constructing \( n \) different instructions \( \{I_{i,1}, I_{i,2}, \dots, I_{i,n}\} \), each containing a distinct type of conflict. Based on the conflict distribution in Table \ref{table.data_statistics}, we generate a total of 864 instructions, each containing a single conflict. 
We denote the sets of instructions containing specific conflict types as \( \mathcal{I}_{CC}, \mathcal{I}_{KK}, \dots, \mathcal{I}_{KP} \), where \textbf{CC}, \textbf{KK}, and \textbf{KP} refer to the conflict types defined earlier. 
We then combine \( \mathcal{I}_0 \) with the conflicting instructions to form nine distinct experiment subsets:
\[
\mathcal{S}_{CC} = \mathcal{I}_0 \cup \mathcal{I}_{CC}, \quad 
\dots, \quad 
\mathcal{S}_{KP} = \mathcal{I}_0 \cup \mathcal{I}_{KP}.
\]
Each subset consists of 100 conflict-free instructions (\( \mathcal{I}_0 \)) and a balanced number of instructions containing a single conflict. 
The subset sizes are as follows: \( \mathcal{S}_{CC}, \mathcal{S}_{KK}, \mathcal{S}_{PP}, \mathcal{S}_{LL}, \mathcal{S}_{FF}, \mathcal{S}_{SS}, \mathcal{S}_{KP} \) each contain 200 instructions, while \( \mathcal{S}_{PC} \) and \( \mathcal{S}_{KP} \) contain 194 and 170 instructions, respectively.

\begin{table*}[!t]
    \footnotesize
    \centering
    % \resizebox{1\textwidth}{!}{%
    % \setlength{\tabcolsep}{1.5pt}% column space
    % \begin{tabular}{ll|cccccc|ccc|ccc}
    \begin{tabular}{
    m{0.01\textwidth}<{\raggedright}
    m{0.26\textwidth}<{\raggedright}
    m{0.025\textwidth}<{\centering}
    m{0.025\textwidth}<{\centering}
    m{0.025\textwidth}<{\centering}
    m{0.025\textwidth}<{\centering}
    m{0.025\textwidth}<{\centering}
    m{0.025\textwidth}<{\centering}|
    m{0.025\textwidth}<{\centering}
    m{0.025\textwidth}<{\centering}
    m{0.025\textwidth}<{\centering}|
    m{0.055\textwidth}<{\centering}
    m{0.055\textwidth}<{\centering}
    m{0.055\textwidth}<{\centering}
    }
    \toprule
    &\textbf{Models }&\textbf{CC}&\textbf{KK}&\textbf{PP}&\textbf{LL}&\textbf{FF}  &\textbf{SS}&\textbf{KP}&\textbf{PC}&\textbf{PS}&\textbf{IntraA}&\textbf{InterA}&\textbf{Average}\\
    % \midrule
    \hline
    &Random Guess & 50.0 & 50.0 & 50.0 & 50.0 & 50.0 & 50.0 & 50.0 & 49.2 & 45.2 & 50.0 & 48.1 & 49.4\\
    % &DeBERTa-XLarge-MNLI ($C>E$) & 86.6 & 85.1 & 83.0 & 62.1 & 85.6 & 86.1 & 79.8 & 84.3 & 57.7 & 81.4 & 73.9 & 78.9 \\
    % &DeBERTa-XLarge-MNLI (Max) & 98.0 & 89.5 & 89.5 & 50.7 & 87.0 & 90.1 & 63.9 & 81.0 & 20.5 & 84.1 & 55.2 & 74.5 \\
    % \midrule
    \hline
    \multirow{7}{*}{\rotatebox{90}{\textbf{Proprietary}}}&GPT-4o (2024-11-20) & 91.9 & 91.3 & 88.7 & 88.1 & 79.8 & 89.8 & 75.1 & 83.7 & 76.1 & 88.3 & 78.3 & 84.9 \\
    &GPT-4o-mini (2024-07-18) & 87.7 & 86.2 & 87.2 & 84.2 & 83.6 & 86.7 & 76.9 & 83.8 & 75.9 & 85.9 & 78.9 & 83.6 \\
    &Claude-4.5-Sonnet (2025-09-29) & 88.5 & 88.5 & 88.5 & 86.0 & 86.5 & 88.5 & \underline{88.5} & 86.8 & \underline{83.6} & 87.7 & \underline{86.3} & \underline{87.3} \\
    &Claude-3.5-Sonnet (2024-10-22) & \textbf{95.7} & \underline{93.1} & \underline{93.1} & \underline{90.5} & \textbf{90.5} & \underline{93.1} & 60.3 & \textbf{89.8} & 73.6 & \textbf{92.7} & 74.6 & 86.6 \\
    &Claude-3.5-Haiku (2024-10-22) & 92.5 & 88.2 & 91.9 & 85.4 & 81.9 & 92.5 & 70.5 & 85.1 & 77.4 & \underline{88.7} & 77.6 & 85.0 \\
    &Gemini-1.5-Pro-Latest & 73.5 & 72.6 & 73.5 & 73.5 & 72.6 & 73.5 & 73.1 & 71.3 & 65.4 & 73.2 & 69.9 & 72.1 \\
    &Gemini-1.5-Flash-Latest & 68.7 & 67.8 & 68.7 & 68.3 & 67.8 & 68.3 & 67.4 & 66.4 & 60.6 & 68.3 & 64.8 & 67.1 \\
    % \midrule
    \hline
    \multirow{11}{*}{\rotatebox{90}{\textbf{Open-source}}}& DeepSeek-R1 (2025-05-28) & \underline{93.1} & \textbf{94.1} & \textbf{93.6} & \textbf{93.1} & \underline{88.1} & \textbf{94.1} & \textbf{93.1} & \underline{89.1} & \textbf{85.3} & \textbf{92.7} & \textbf{89.2} & \textbf{91.5} \\
    &Meta-Llama-3.2-1B-Instruct & 38.7 & 28.8 & 28.8 & 33.3 & 32.2 & 34.4 & 34.4 & 26.3 & 39.0 & 32.7 & 33.2 & 32.9 \\
    &Meta-Llama-3.2-3B-Instruct & 49.5 & 46.3 & 46.3 & 36.9 & 38.7 & 39.6 & 41.3 & 52.6 & 43.2 & 42.9 & 45.7 & 43.8 \\
    &Meta-Llama-3.1-8B-Instruct & 70.9 & 68.3 & 68.3 & 63.3 & 65.6 & 66.7 & 62.7 & 68.9 & 58.8 & 67.2 & 63.5 & 65.9 \\
    &Ministral-8B-Instruct-2410 & 67.9 & 69.3 & 69.3 & 66.9 & 65.0 & 68.3 & 67.9 & 67.9 & 58.4 & 67.8 & 64.7 & 66.8 \\
    &Qwen2.5-0.5B-Instruct & 36.2 & 42.2 & 43.1 & 48.6 & 47.7 & 41.2 & 43.1 & 32.2 & 42.2 & 43.2 & 39.2 & 41.8 \\
    &Qwen2.5-1.5B-Instruct & 36.5 & 37.6 & 33.1 & 24.6 & 33.1 & 33.1 & 31.9 & 35.7 & 29.9 & 33.0 & 32.5 & 32.8 \\
    &Qwen2.5-3B-Instruct & 59.7 & 56.1 & 54.6 & 45.9 & 50.0 & 48.4 & 54.6 & 62.9 & 46.9 & 52.5 & 54.8 & 53.2 \\
    &Qwen2.5-7B-Instruct & 76.3 & 65.4 & 62.8 & 61.9 & 44.9 & 59.2 & 41.5 & 66.2 & 42.9 & 61.8 & 50.2 & 57.9 \\
    &Qwen2.5-14B-Instruct & 90.2 & 80.2 & 79.5 & 79.5 & 65.8 & 81.6 & 57.7 & 83.7 & 64.3 & 79.5 & 68.6 & 75.8 \\
    &Qwen2.5-32B-Instruct & \underline{93.8} & 89.1 & 83.4 & 78.6 & 63.1 & 85.4 & 39.4 & 79.2 & 54.5 & 82.2 & 57.7 & 74.1 \\
    \bottomrule
    \end{tabular}%
    % }
    % \vspace{-3mm}
    \caption{Conflict detection results (\%) of LLMs on different subsets, each containing instructions with a single type of conflict. Here, \textit{conflict types} refer to subsets that contain the corresponding conflict, e.g., \textit{CC} denotes \( \mathcal{S}_{CC} \). 
    `IntraA' and `InterA' denote the average performance across subsets of intra-constraint and inter-constraint conflicts, respectively. The reported metric is the F1-score (F1). The top two results among LLMs are highlighted in \textbf{bold} and \underline{underlined}, respectively. Parenthesized numbers indicate specific dated snapshots of proprietary LLMs.}
    % \vspace{-4mm}
    \label{table.conflict_detection}
\end{table*}

\paragraph{Instructions with Multiple Conflicts.}\label{section.multiple_conflicts}
To construct instructions with \( k \) constraints (\( k \in \{1, 2, 3, 4, 5, 6 \}\)), for each instruction \( I_i \in \mathcal{I}_0 \), we randomly select \( k \) conflicts from its corresponding conflict set \( \{c_1, c_2, \dots, c_n\} \), shuffle them, and append them to \( I_i \). Due to computational constraints, we generate a single instruction with \( k \) conflicts for each \( I_i \).
This process results in the set \( \mathcal{I}_k \), which contains 100 instructions, each with \( k \) conflicts.

We will evaluate LLM performance on conflict detection and conflict resolution across these subsets.

\subsection{Evaluation Models}
We evaluate a range of models for conflict detection and resolution, categorizing them into two primary groups: (1) Seven Proprietary LLMs, including GPT-4o (gpt-4o-2024-11-20), GPT-4o-mini (gpt-4o-mini-2024-07-18) \cite{Achiam2023GPT4TR}, Claude-4.5-Sonnet (claude-4-5-sonnet-2025-09-29), Claude-3.5-Sonnet (claude-3-5-sonnet-20240620), Claude-3.5-Haiku (claude-3-5-haiku-20240620) \cite{Anthropic}, Gemini-1.5-Pro-Latest (gemini-1.5-pro-latest), and Gemini-1.5-Flash-Latest (gemini-1.5-flash-latest)\footnote{We used the Gemini-1.5 API in January 2025.} \cite{Reid2024Gemini1U}. (2) Eleven Open-source LLMs, including DeepSeek-R1 (2025-05-28) \cite{guo2025deepseek}, Meta-Llama-3.2-1B-Instruct, Meta-Llama-3.2-3B-Instruct, Meta-Llama-3.1-8B-Instruct \cite{dubey2024llama}, Mistral-8B-Instruct-2410 \cite{Mistral}, and Qwen2.5-[0.5, 1.5, 3, 7, 14, 32]B-Instruct \cite{qwen2.5}. 
For all models, we set the maximum output length to 2048 tokens and use a temperature of 0 to ensure deterministic outputs. 
Experiments involving open-source LLMs were conducted using A100 GPUs with 40GB of memory, while proprietary LLMs were accessed through their official APIs.

\section{Experiment Results on Conflict Detection}
In this section, we explore the conflict detection task, which evaluates whether LLMs can identify conflicting instructions. Given an instruction \( I \), the conflict detection task is formulated as a function \( f(I) \in \{\text{Yes, No}\} \), where \( f \) can be instantiated by an LLM. The prompt used for conflict detection is provided in Table 6 of Appendix E.

\subsection{Instructions with a Single Conflict} 
Table \ref{table.conflict_detection} shows the conflict detection performance of various models. Our key findings are:\\
(1) \textbf{Proprietary LLMs exhibit superior performance in conflict detection, with the Claude family of models being particularly dominant.} This is evidenced by their notably high F1 scores—Claude-4.5-Sonnet (87.3\%), Claude-3.5-Sonnet (86.6\%), and Claude-3.5-Haiku (85.0\%). \\
(2) \textbf{Open-source models with fewer than 7B parameters struggle with conflict detection.}  
Models such as Meta-Llama-3.2-3B-Instruct and Qwen2.5-1.5B-Instruct underperform relative to random guessing across most conflict types, indicating their inability to detect conflicts effectively.\\
(3) \textbf{Detecting intra-constraint conflicts is easier than inter-constraint conflicts.} For instance, Claude-3.5-Sonnet scores 92.7\% on intra-constraint conflict subsets but only 74.6\% on inter-constraint conflict subsets. This pattern is consistent with other strong models, suggesting that intra-constraint conflicts are more recognizable than inter-constraint conflicts.
 
These findings highlight the strength of proprietary LLMs in conflict detection and the challenges faced by smaller open-source models.

\subsection{Instructions with Multiple Conflicts}
\begin{figure}
    \centering
    \includegraphics[width=0.45\textwidth]{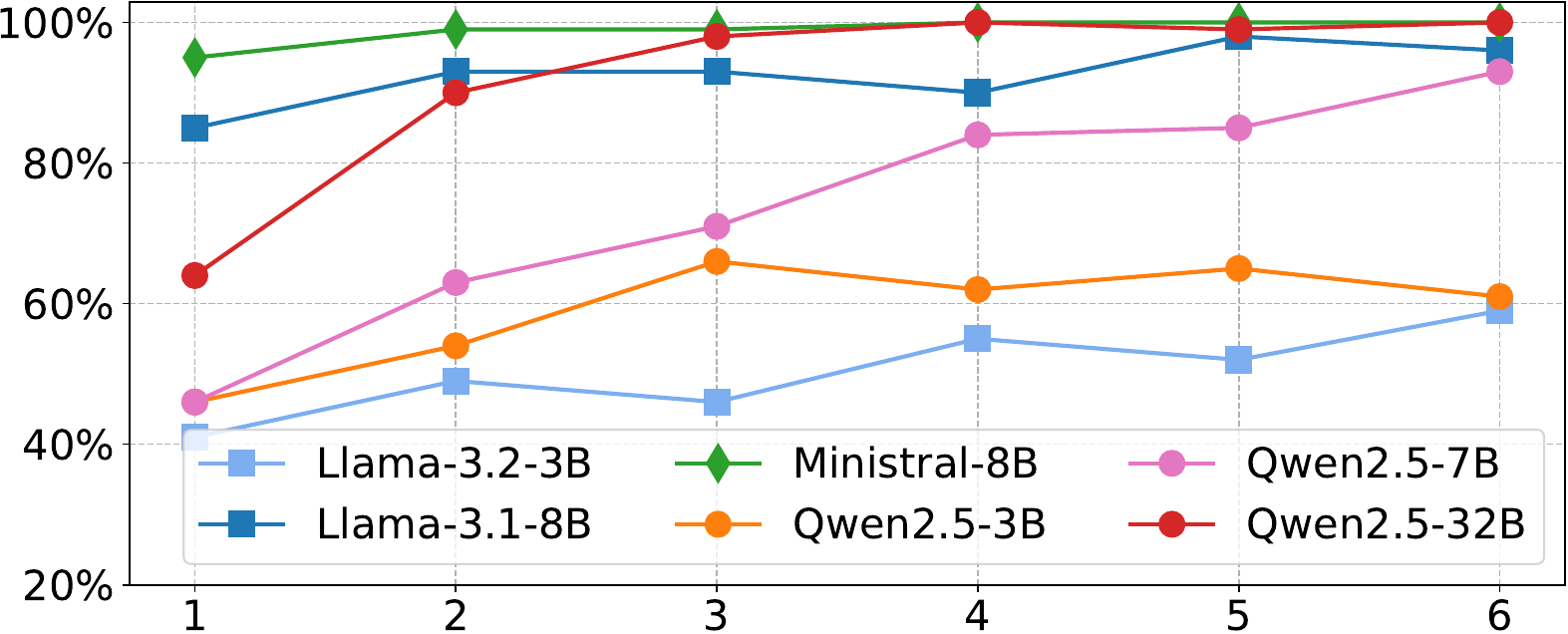} 
    % \vspace{-2mm}
    \caption{ Conflict detection results of LLMs for instructions with varying numbers of conflicts (i.e., instructions in \( \mathcal{I}_k \)). 
    The x-axis denotes the number of conflicts per instruction. 
    The reported metric is Recall.
    }
    \label{fig.conflict_detection_density}
    % \vspace{-4mm}
\end{figure}

Figure \ref{fig.conflict_detection_density} illustrates the conflict detection performance of various LLMs as the number of conflicts in instructions increases. \textbf{As the number of conflicts within an instruction grows, larger models generally exhibit improved detection performance.} This trend is particularly evident in Qwen2.5-[7, 32]B. 
However, smaller open-source models still struggle with conflict detection. 
Even when instructions contain multiple conflicts, models with fewer than 7B parameters, such as LLaMA-3.2-3B and Qwen2.5-3B, exhibit lower recall in identifying conflicts. This suggests that smaller models may lack the necessary reasoning capacity to detect conflicting constraints.

\section{Experiment Results on Conflict Resolution}
% 前面的章节现实LLMs especially, Proprietary LLMs ex-
% hibit strong conflict detection capabilities. 
% Given this, In this section, we will further examine how LLMs handle instructions with conflicting constraints, whether they can explicitly acknowledge the existence of these conflicting constraints in their responses. 
% % simulating real-world scenarios where user instructions contain mutually contradictory requirements. 
% To be specific, We first observe LLMs' behaviors in response to such conflicts, and then analyze the effect of conflicting constraints on the original conflict-free constraints.

In the previous section, we demonstrated that LLMs, particularly proprietary ones, exhibit strong conflict detection capabilities when faced with conflicting instructions. Building on this finding, this section further investigates how LLMs handle instructions with conflicting constraints, and whether they can respond safely—that is, by explicitly acknowledging conflicts in their responses and notifying users. 
Specifically, we first observe LLMs' behaviors in response to such conflicts, and then analyze the effect of conflicting constraints on the original conflict-free constraints.

\subsection{Analysis on Conflict Resolution Behaviors}
\paragraph{Typical Conflict Resolution Behaviors.} 
In Section \ref{section.multiple_conflicts}, we create six subsets \( \mathcal{I}_k \), where each instruction contains \( k \) conflicts. We feed these conflicting instructions into LLMs and analyze their responses, classifying their behaviors into four types. 
1. \textbf{Conflict Unacknowledged}: The model does not indicate the presence of conflicts in its response and directly provides a response to the instruction. 
2. \textbf{Conflict Acknowledged, Clarification Requested}: The model recognizes that the instruction contains conflicts, refuses to respond, and explicitly asks users for clarification. 
3. \textbf{Conflict Acknowledged, Autonomously Resolved}: The model identifies conflicts, resolves them on its own, and provides a response to the resolved instruction. 
4. \textbf{Other Behaviors}: The model refuses to respond for reasons unrelated to conflicts.

The first behavior is particularly problematic, as the model fails to inform users of conflicts while generating a response that satisfies only a subset of constraints. This may mislead users into accepting incomplete or incorrect responses without realizing that their instruction contains conflicts. 
In contrast, Behaviors 2 and 3 explicitly acknowledge the conflicts. Behavior 3 autonomously resolves them, while Behavior 2 seeks clarification from users. Among these, Behavior 2 is the most desirable, as it ensures transparency and allows users to control the conflict resolution process.

\begin{figure*}
    \centering
    \includegraphics[width=0.9\textwidth]{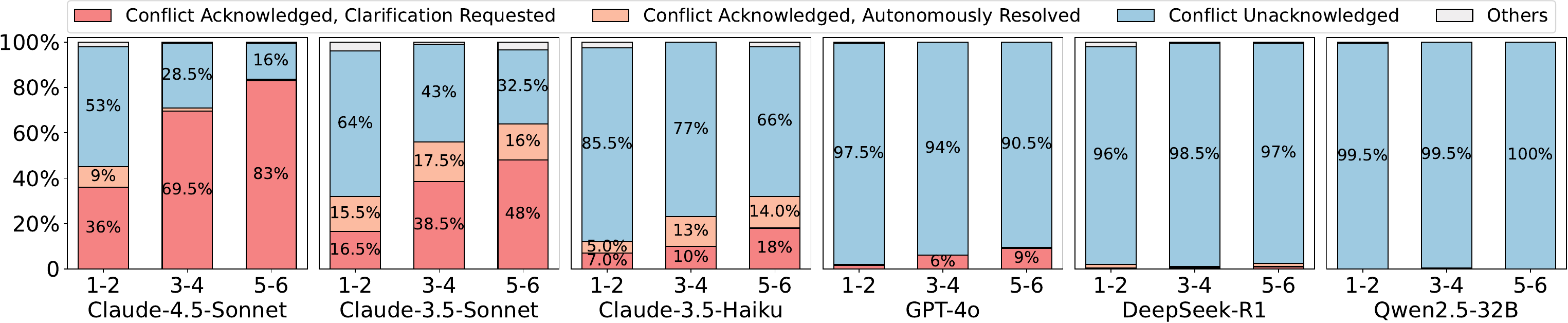} 
    % \vspace{-7mm}
    \caption{ 
    Distributions of conflict resolution behaviors exhibited by different LLMs when responding to instructions with varying numbers of conflicts. 
    The x-axis denotes the number of conflicts per instruction.
    }
    \label{fig.conflict_resolution_density}
    % \vspace{-4mm}
\end{figure*}

\paragraph{Distribution of Conflict Resolution Behaviors.} 
To systematically analyze LLM behavior, we use GPT-4o to assign behavior labels to 3,600 responses from six LLMs (see Table 7 in Appendix E for the evaluation prompt). 
To check the quality of GPT-4o's assessment, we manually annotate behavior labels for 400 responses, achieving 98\% agreement with GPT-4o's judgments. Besides GPT-4o, we employ Gemini-2.5-Pro as another judge. The Cohen's Kappa between both LLMs is 0.746, showing substantial agreement ($>0.6$) \cite{Landis1977TheMO} (Please refer to Appendix F for more details). 
Figure \ref{fig.conflict_resolution_density} presents the distribution of conflict resolution behaviors exhibited by different LLMs when responding to instructions with varying numbers of conflicts. We summarize the key findings as follows:\\
\textbf{(1) GPT-4o, DeepSeek-R1, and Qwen2.5-32B predominantly exhibit Behavior 1}: 
% In most cases, both GPT-4o and Qwen2.5-32B directly respond to instructions without explicitly acknowledging the presence of conflicts. 
However, this does not imply that these LLMs lack the ability to detect conflicts. 
% Interestingly, this behavior is not necessarily due to an inability to detect conflicts. 
% As evidenced in Table \ref{table.conflict_detection}, GPT-4o achieves an average F1 score of 84.9 in conflict detection when instructions contain a single conflict. 
In Figure \ref{fig.conflict_detection_density}, Qwen2.5-32B can identify conflicts with near 100\% accuracy when more than two conflicts are present in an instruction. \textbf{Despite their conflict detection capabilities, they fail to explicitly acknowledge conflicts in most cases.}\\
\textbf{(2) Claude models exhibit conflict-aware behavior that scales with the number of conflicts.} 
In Claude-3.5-Sonnet, the combined proportion of Behaviors 2 and 3 increases from 32.0\% when handling instructions with 1-2 conflicts to 64.0\% when handling instructions with 5-6 conflicts. 
A similar trend is observed in Claude-4.5-Sonnet and Claude-3.5-Haiku.
% where a higher number of conflicts leads to Behaviors 2 and 3. 
However, despite the presence of multiple conflicts in instructions, Behavior 1 still constitutes a significant proportion of Claude models' responses.

These findings underscore the need to improve LLMs to adopt safe conflict resolution behaviors when faced with conflicts, which is essential to ensure reliable responses.

% Enhancing models' transparency in conflict resolution is essential for ensuring reliable and interpretable AI responses.

\begin{table}[!h]
    \scriptsize
    \centering
    \setlength{\tabcolsep}{4pt}% column space
    % \resizebox{0.48\textwidth}{!}{
    \begin{tabular}{lc|cccc|cc|c}
    % \toprule
    \hline
    \multirow{2}{*}{\textbf{Model}}& \multirow{2}{*}{\textbf{GP}} &\multicolumn{4}{c}{\textbf{ $\mathcal{I}_1$}}&\multicolumn{2}{c}{\textbf{ $\mathcal{I}_0$}}&\multirow{2}{*}{\textbf{F1}}\\
    % \cmidrule(lr){3-6} \cmidrule(lr){7-8}
    \cline{3-6} \cline{7-8}
    &&\textbf{B1$\downarrow$}&\textbf{B2$\uparrow$}&\textbf{B3}&\textbf{B4}&\textbf{B5$\uparrow$}&\textbf{B6$\downarrow$}&\\
    % \midrule
    \hline
    \multirow{2}{*}{GPT-4o} 
                            &\redcross & 97 & 1 & 1 & 1 & 100 & 0 &-\\
                            & \greencheck  & \textbf{4} &  \textbf{96}  &0 & 0& \underline{60} & \underline{40} &81.4\\
    % \midrule
    \hline
    \multirow{2}{*}{GPT-4o-mini} 
                                & \redcross& 98& 0&0 & 2&100 & 0 &-\\
                                & \greencheck& \textbf{4} & \textbf{96} &0 & 0& \underline{47} & \underline{53} &77.1\\
    % \midrule
    \hline
    \multirow{2}{*}{Claude-3.5--Sonnet} &\redcross & 78& 7& 11& 4& 100& 0 &-\\
    & \greencheck& \textbf{18} & \textbf{82} &0 & 0& \underline{72} & \underline{28} &78.1\\

    \hline
    \multirow{2}{*}{Claude-3.5-Haiku} &\redcross & 93& 2& 2& 3& 100& 0 &-\\
    & \greencheck& \textbf{16} & \textbf{84} &0 & 0& \underline{78} & \underline{22} &81.6\\
    % \bottomrule
    \hline
    \end{tabular}%
    % }
    % \vspace{-2mm}
    \caption{
    Distribution (\%) of LLM behaviors with or without the guiding prompt (GP) designed to detect and resolve instruction conflicts using Behavior 2. Here, B refers to behavior types. $\mathcal{I}_1$ and $\mathcal{I}_0$ represent instructions with one conflict and without conflicts, respectively. For conflict-free instructions ($\mathcal{I}_0$), we report two types of model behaviors: \textbf{Behavior 5} (LLMs determine that the instruction has no conflict and executes it directly) and \textbf{Behavior 6} (LLMs incorrectly detect conflicts and unnecessarily  asks for clarification). F1 denotes the F1 score of LLMs in identifying instruction conflicts when using the GP. Results highlighted in \textbf{bold} and \underline{underlined} indicate whether the behavioral changes meet or fail to meet expectations, respectively.
    }
    % \vspace{-4mm}
    \label{table.conflict_resolution_behavior}
\end{table}

\subsection{Prompting LLMs to Resolve Instruction Conflicts Using Desired Behaviors}
LLMs often fail to explicitly acknowledge conflicting instructions. This study investigates whether prompt engineering can guide LLMs to identify and resolve such conflicts according to desired behavioral patterns. To explore this, we prepend user instructions with the prompt designed to detect and resolve instruction conflicts with Behavior 2, as detailed in Table 8 of Appendix E. 
As shown in Table \ref{table.conflict_resolution_behavior}, this prompt can effectively induce LLMs to adopt the predefined desired Behavior 2 (acknowledging conflict and requesting clarification). However, it also causes LLMs to behave overly conservatively, asking for clarification even when no conflict exists (Behavior 6), thereby degrading the user experience. These findings suggest that \textbf{while prompt engineering can influence conflict resolution behavior, it alone is insufficient for achieving both desired conflict resolution and accurate execution of non-conflicting instructions.}

\begin{figure*}[h!]
% \vspace{-2mm}
\begin{minipage}{0.31\textwidth}
    \centering
    \includegraphics[width=1\linewidth]{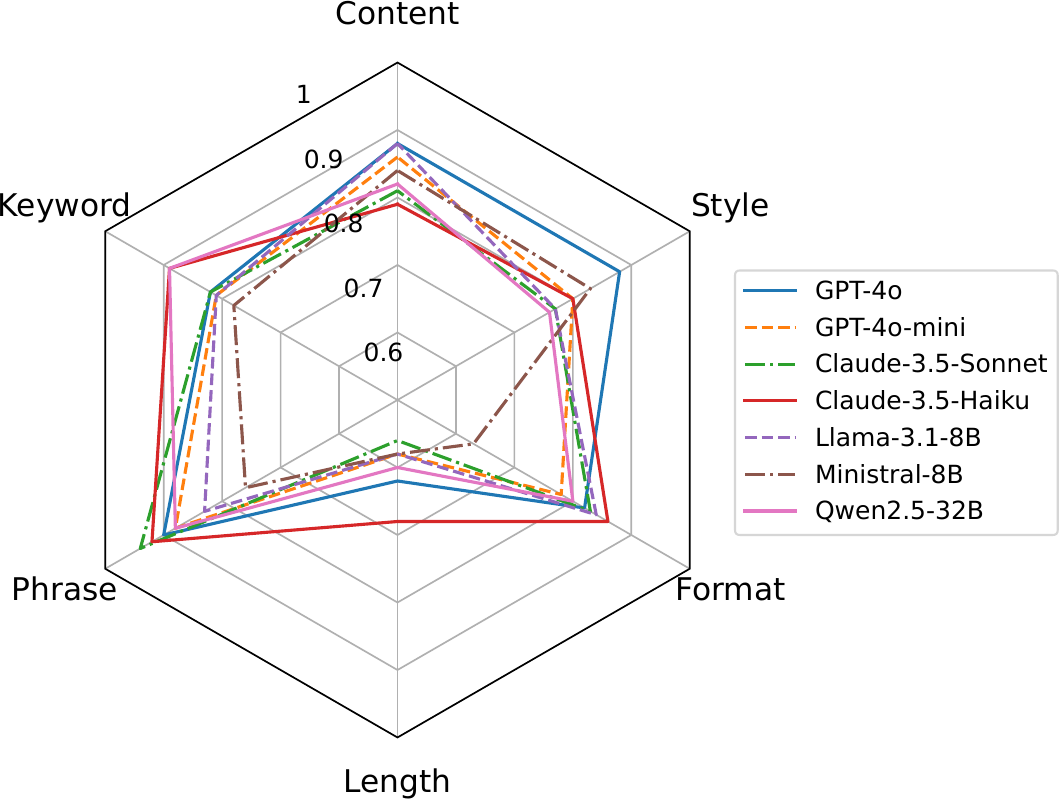}
    % \vspace{-7mm}
    \caption{ CSR results of various LLMs across different constraint types.
  }
  \label{fig.instruction_following}
\end{minipage}
\hfill
\begin{minipage}{0.65\textwidth} 
    \centering
        \subfigure[Constraint satisfaction rates of OC]{
      \centering
      \includegraphics[width=0.53\textwidth]{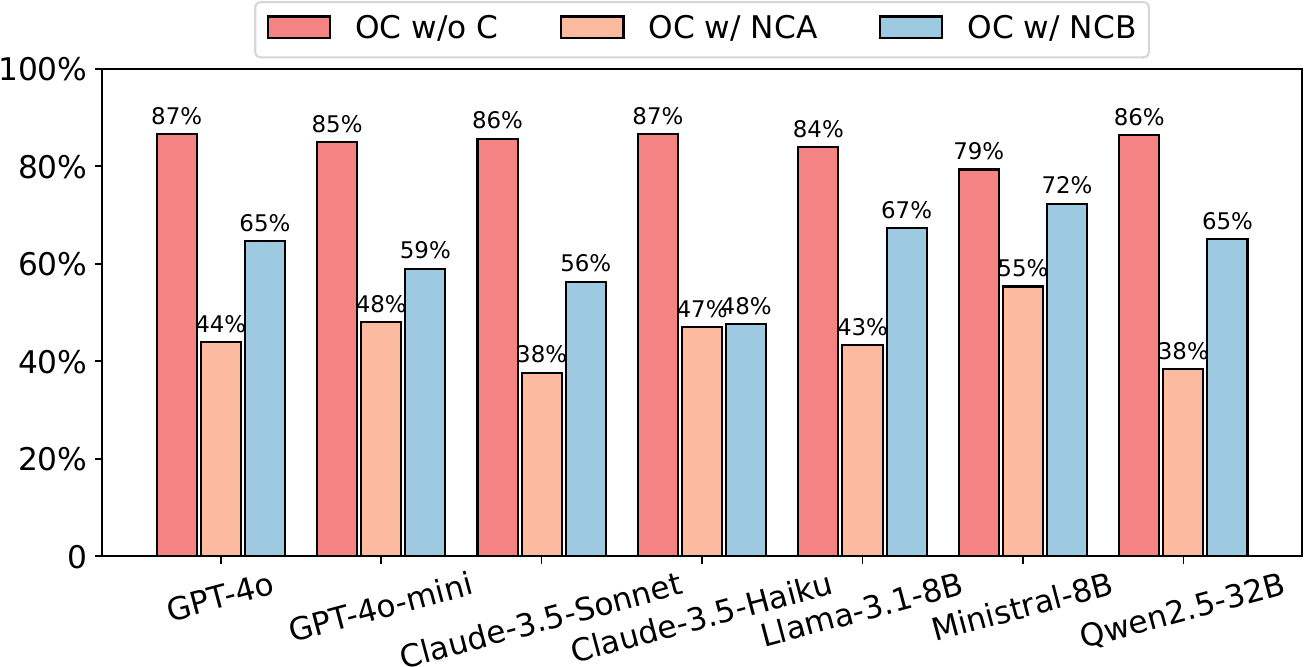} 
    }
    \hspace{-3mm}
    \subfigure[Constraint satisfaction rates of NC]{
      \centering
      \includegraphics[width=0.44\textwidth]{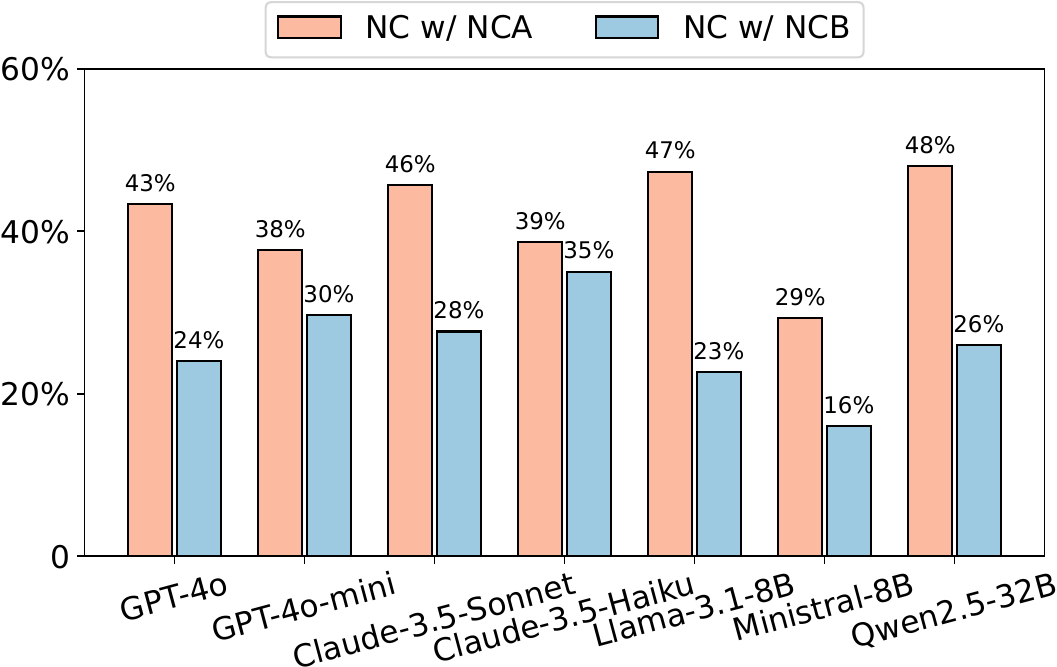} 
    }
    % \vspace{-5mm}
    \caption{The impact of the order of NC on the constraint satisfaction rates of both OC and NC. `w/o C' denotes the absence of conflicts, while `NCA' and `NCB' indicate that NC appears after and before OC, respectively.}
   \label{fig.conflict_order}
\end{minipage} 
% \vspace{-6mm}
\end{figure*}

\subsection{Analysis on Constraint Priority}

\paragraph{Constraint-Following Ability of LLMs on Conflict-Free Instructions.} 
We first feed each conflict-free instruction \( I_i \in \mathcal{I}_0 \) into LLMs and evaluate their \textbf{Constraint Satisfaction Rate (CSR)} in the absence of conflicting constraints. CSR is defined as follows: 
\begin{eqnarray}\label{eq:1}
  CSR =  \frac{1}{M}\sum_{i=1}^{N}\sum_{j=1}^{l_i}I_i^j,
\end{eqnarray}
where \( I_i^j = 1 \) if the \( j \)-th constraint of the \( i \)-th instruction is satisfied and \( I_i^j = 0 \) otherwise. Here, \( l_i \) denotes the number of constraints in \( I_i \), \( N \) represents the number of instructions, and \( M \) is the total number of constraints in all instructions.  

We use GPT-4o to evaluate whether the model's output satisfies the specified constraints (Table 9 in Appendix E shows the evaluation prompt). To assess the evaluation quality of GPT-4o, we manually labeled 150 constraints and  verified whether each constraint was satisfied in LLMs' responses. 
Table 11 in Appendix F shows that automatic evaluation aligns closely with human judgment.
% The consistency between human and GPT-4o annotations is presented in Table \ref{table.consistency_instruction_following}. The results demonstrate that automatic evaluation aligns closely with human judgment. 
Figure \ref{fig.instruction_following} presents the CSR results of seven LLMs across different constraint types, revealing a clear performance pattern: the CSR score is notably lowest for length constraints but higher for the other five constraint types.

% \begin{figure}
%     \centering
%     \includegraphics[width=0.45\textwidth]{./figures/instruction_following-crop.pdf} 
%     \vspace{-2mm}
%     \caption{ 
%     CSR results of various LLMs across different constraint types. 
%     }
%     \vspace{-4mm}\label{fig.instruction_following}
% \end{figure}

% \begin{figure*}[!ht]
%   \centering
%     \subfigure[Constraint satisfaction rates of OC]{
%       \centering
%       \includegraphics[width=0.52\textwidth]{./figures/conflict_resolution_OC-crop.pdf} 
%     }
%     % \hspace{-1mm}
%     \subfigure[Constraint satisfaction rates of NC]{
%       \centering
%       \includegraphics[width=0.42\textwidth]{./figures/conflict_resolution_NC-crop.pdf} 
%     }
%     \vspace{-4mm}
%       \caption{ 
%         The impact of the order of NC on the constraint satisfaction rates of both OC and NC. `w/o C' denotes the absence of conflicts, while `NCA' and `NCB' indicate that NC appears after and before OC, respectively.
%       }
%       \vspace{-4mm}
%       \label{fig.conflict_order}
% \end{figure*}

\paragraph{Impact of Conflicting Constraints on LLMs' Constraint-Following Ability.}  
As shown in Figure \ref{fig.conflict_resolution_density}, when an instruction contains only a few conflicts, LLMs predominantly exhibit Behavior 1, meaning they tend to satisfy some of the constraints in the instruction. 
To further investigate how \textbf{N}ewly introduced conflicting \textbf{C}onstraints (NC) affect LLMs' ability to adhere to \textbf{O}riginal \textbf{C}onstraints (OC), we focus on instructions containing a single conflict. Given computational constraints, we examine three types of conflicts: CC, KK, and PP. 
To examine the effect of NC's position, we construct two subsets for each conflict type: 
\textbf{NCA subsets} (\(\mathcal{I}_{CC}, \mathcal{I}_{KK}, \mathcal{I}_{PP}\)), where NC is introduced \textbf{after} OC, and \textbf{NCB subsets} (\(\mathcal{I}_{CC}^{\prime}, \mathcal{I}_{KK}^{\prime}, \mathcal{I}_{PP}^{\prime}\)), where NC is introduced \textbf{before} OC.  
% \begin{itemize}
%     \item \textbf{NCA subsets} (\(\mathcal{I}_{CC}, \mathcal{I}_{KK}, \mathcal{I}_{PP}\)): NC is introduced \textbf{after} OC.  
%     \item \textbf{NCB subsets} (\(\mathcal{I}_{CC}^{\prime}, \mathcal{I}_{KK}^{\prime}, \mathcal{I}_{PP}^{\prime}\)): NC is introduced \textbf{before} OC.  
% \end{itemize}  
Each subset contains 100 single-conflict instructions. We input these into LLMs and use GPT-4o to evaluate whether their responses satisfy OC or NC (Table 10 in Appendix E shows the evaluation prompt). To validate the reliability of GPT-4o's assessment, we manually annotated 100 conflicting cases, achieving 90\% agreement with GPT-4o's judgments. 

Figure \ref{fig.conflict_order}(a) shows the impact of NC on LLMs' ability to satisfy OC. The results reveal the following key observations: 
(1) NC significantly reduces OC satisfaction rates, suggesting that newly introduced conflicting constraints degrade previous constraints. 
(2) OC is more likely to be followed when NC appears before OC rather than after it. This suggests that the \textbf{later a constraint appears in an instruction, the more likely it is to be followed}. 
In Figure \ref{fig.conflict_order}(b), NC is more likely to be followed when it appears \textbf{later} (NCA) rather than earlier (NCB), further reinforcing the idea that \textbf{conflicting constraints appearing later in an instruction are more likely to be satisfied}.

\section{Related Work}

% \subsection{Controllable Text Generation}
\textbf{Controllable Text Generation} focuses on guiding language models to generate text with specific attributes, such as sentiment \cite{keskar2019ctrl, Dathathri2020Plug}, 
lexical constraints \cite{he-2021-parallel,he2021show,he-etal-2022-metric}, 
length \cite{kikuchi-etal-2016-controlling, fan-etal-2018-controllable}. Recent studies have constructed data based on these controllable tasks to evaluate \cite{zhou2023instruction, sun-etal-2023-evaluating} or enhance the instruction-following ability of LLMs \cite{pmlr-v202-zhou23g}. 
Unlike these studies, we investigate how LLMs detect and resolve conflicts when given instructions with conflicting constraints, thereby providing new insights into their instruction-following capabilities.

% Unlike prior work, which assumes that all constraints within instructions are consistent, we assess LLMs' ability to detect and resolve conflicting constraints, offering new insights into their behavior when handling instructions with conflicts.

% \subsection{Conflict Detection}
\paragraph{Conflict Detection} has been extensively studied in natural language inference \cite{bowman-etal-2015-large, williams-etal-2018-broad} and fact verification \cite{thorne-etal-2018-fever}, aiming to detect contradictions between two statements or between claims and external evidence sources. More recently, research has expanded to detecting conflicts among retrieved documents \cite{jiayang-etal-2024-econ}, or discrepancies between LLMs' parametric knowledge and retrieved documents \cite{chen-etal-2022-rich,neeman-etal-2023-disentqa, xie2024adaptive}. 
Meanwhile, hallucination detection in LLMs \cite{manakul-etal-2023-selfcheckgpt, min-etal-2023-factscore} investigates false  content generated by LLMs. 
While these studies explore different aspects of conflict detection, they do not focus on conflicting instructions where multiple constraints contradict each other. Our work extends beyond these domains by systematically evaluating how LLMs detect and resolve explicit conflicts within user instructions.

\section{Conclusion}
We introduce ConInstruct, a benchmark designed to evaluate LLMs’ ability to detect and resolve conflicting constraints within instructions. Our findings reveal that while proprietary LLMs demonstrate strong conflict detection capabilities, they often fail to explicitly communicate conflicts to users, instead generating responses that only partially satisfy the given constraints. This highlights a critical gap in instruction-following: despite recognizing conflicts, LLMs struggle to transparently convey them. Future research should focus on enhancing LLMs’ ability to explicitly notify users of conflicts and seek clarification, improving their reliability in real-world applications that demand precise adherence to instructions.

\section{Acknowledgments}
This work is supported by HKU-SCF FinTech Academy, Shenzhen-Hong Kong-Macao Science and Technology Plan Project (Category C Project: SGDX20210823103537030), and Theme-based Research Scheme of RGC, Hong Kong (T35-710/20-R). Linlin Yu did not receive any financial support for this work and contributed only by developing the research ideas, participating in discussions, and providing feedback on the manuscript.

\bibliography{aaai2026}

\appendix
% \clearpage
\section{Limitations and Future Work}
\paragraph{Limitations.} Despite the insights provided by ConInstruct, our study has several limitations. While our benchmark covers a diverse range of constraints and conflicts, it may not fully encompass all possible forms of instruction inconsistencies. In designing conflicting constraints, we prioritized the feasibility of evaluating constraint satisfaction using LLMs or automated programs. This consideration led us to avoid overly complex constraints and ambiguous conflicts. 

Another limitation is that our dataset primarily focuses on text-based instructions, restricting its applicability to multimodal scenarios where conflicts may arise in image, audio, or video-based instructions. 

\paragraph{Future Work.} Future work should expand the benchmark to include more challenging forms of instruction conflicts, such as implicit conflicts that require multi-hop reasoning or commonsense inference, as well as conflicts that emerge in multimodal settings. Such extensions would provide a more comprehensive evaluation of LLMs’ conflict detection and resolution capabilities.

% \clearpage
% \appendix

\section{Data Distribution}
\begin{figure}[h]
    \centering
    \includegraphics[width=0.48\textwidth]{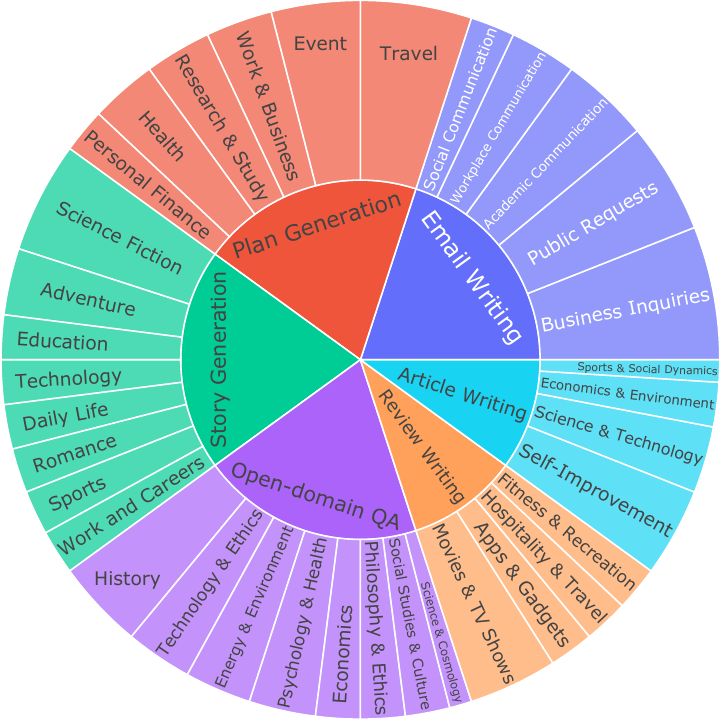} 
    % \vspace{-3mm}
    \caption{ 
    Task and domain distribution of ConInstruct. The size of each sector reflects its proportion in the dataset.
    }
    % \vspace{-4mm}
    \label{fig.data_distribution}
\end{figure}

The distribution of tasks and domains across the ConInstruct dataset is illustrated in Figure \ref{fig.data_distribution}.

\section{Constraint Dimensions}\label{section.constraint_dimensions}

\subsection{Content Constraints} 
Content constraints involve incorporating specific details related to the content. These details may encompass various aspects, such as reasons, purposes, topics, background information, budgets, targets, and more.

\subsection{Keyword Constraints}
\paragraph{1. Keywords Inclusion:} This constraint specifies the inclusion of specific keywords. Examples of this constraint:
\begin{itemize}
    \item Include all of the following keywords in the response: \{keyword list\}.
    \item Include at least/at most/exactly \{N\} of the following keywords in the response: {keyword list}.
    \item Include either \{keyword1\} or \{keyword2\}, but not both, in the response.
\end{itemize}

\paragraph{2. Forbidden Words:} This constraint specifies keywords that must not be included. Examples of this constraint:
\begin{itemize}
    \item  Do not include the following forbidden keywords in the response: \{forbidden keyword list\}.
\end{itemize}

\paragraph{3. Keyword Frequency:} This constraint specifies the frequency of keywords. Examples of this constraint:
\begin{itemize}
    \item The keyword \{keyword\} must appear at least/at most/exactly \{N\} times in the response.
\end{itemize}

\paragraph{4. Letter Frequency:} This constraint specifies the frequency of letters. Examples of this constraint:
\begin{itemize}
    \item The letter \{letter\} must appear at least/at most/exactly \{N\} times in the response.
\end{itemize}

\paragraph{5. Keyword Order:} This constraint specifies the order of keywords. Examples of this constraint:
\begin{itemize}
    \item The keyword \{keyword1\} must appear before/after the keyword \{keyword2\} in the response.
\end{itemize}

\paragraph{6. Keyword Proximity:} This constraint specifies the distance between keywords. Examples of this constraint:
\begin{itemize}
    \item The keyword \{keyword1\} must appear at least/at most/exactly \{N\} words/sentences/paragraphs away from the keyword \{keyword2\}.
    \item The keywords \{keyword list\} must/must not appear in the same paragraph/sentence.
\end{itemize}

\paragraph{7. Keyword Position:} This constraint specifies the positions of keywords. Examples of this constraint:
\begin{itemize}
    \item The keywords \{keyword list\} must/must not appear in the first/last/n-th paragraph/sentence of the response.
\end{itemize}

\paragraph{8. Keyword Part-of-speech:} This constraint specifies the part-of-speech tag for keywords, which possess multiple part-of-speech tags in the Oxford Dictionary. Do not apply this constraint to keywords with only a single part-of-speech tag. Examples of this constraint:
\begin{itemize}
    \item The keyword \{keyword\} must appear in the response and used as \{part-of-speech tag in the Oxford Dictionary\}.
\end{itemize}

\paragraph{9. Keyword Definition:} This constraint specifies the definition for keywords, which possess multiple definitions in the Oxford Dictionary. This constraint can only be used to verbs or adjectives. Do not apply this constraint to keywords with fewer than three definitions. Examples of this constraint:
\begin{itemize}
    \item The keyword \{verb or adjective\} must appear in the response and convey the specified definition \{definition in the Oxford Dictionary\}.
\end{itemize}

\subsection{Phrase Constraints}
\paragraph{1. Phrase Inclusion:} This constraint specifies the inclusion of specific phrases.  The specific phrase must contain at least four words. Examples of this constraint:
\begin{itemize}
    \item Include the phrase \{phrase\} in the response.
\end{itemize}

\paragraph{2. Phrase Frequency:} This constraint specifies the frequency of phrases. The specific phrase must contain at least four words. Examples of this constraint:
\begin{itemize}
    \item The phrase \{phrase\} must appear at least/at most/exactly \{N\} times in the response.
\end{itemize}

\paragraph{3. Phrase Position:} This constraint specifies the positions of keywords. The specific phrase must contain at least four words. Examples of this constraint:
\begin{itemize}
    \item Start/Finish the response/n-th paragraph with the phrase \{phrase\}. 
    \item Include the phrase \{phrase\} in n-th paragraph. 
\end{itemize}
\subsection{Length Constraints}
\paragraph{1. Number of Paragraphs:} This constraint specifies the required number of paragraphs in the response. Examples of this constraint:
\begin{itemize}
    \item The response must contain at least/at most/exactly \{N\} paragraphs.
\end{itemize}

\paragraph{2. Number of Sentences:} This constraint specifies the required number of sentences in the response or within specific paragraphs. Examples of this constraint:
\begin{itemize}
    \item The response must contain at least/at most/exactly \{N\} sentences.
    \item The n-th paragraph must contain at least/at most/exactly \{N\} sentences.
    \item Each paragraph must contain at least/at most/exactly \{N\} sentences.
\end{itemize}

\paragraph{3. Number of Words:} This constraint specifies the required number of words in the response, or within specific paragraphs or sentences. Examples of this constraint:
\begin{itemize}
    \item The response must contain at least/at most/exactly \{N\} words.
    \item The n-th paragraph/sentence must contain at least/at most/exactly \{N\} words.
    \item Each paragraph/sentence must contain at least/at most/exactly \{N\} words.
\end{itemize}

\subsection{Format Constraints}
\paragraph{1. JSON Format:} This constraint requires the entire response to be wrapped in JSON format and follow specific JSON structure.
\begin{itemize}
    \item The response must include the following keys: \{key list\}.
    \item The response must include at least/at most/exactly \{N\} of the following types: Number, String, Boolean, Array, or Object.
    \item The value of the key \{key\} must be a Number/String/Boolean/Array/Object.
    \item The value of the key \{key\} must be an integer equal to/less than/greater than \{N\}.
    \item The value of the key \{key\} must be an Array/Object containing at least/at most/exactly \{N\} elements.
\end{itemize}

\paragraph{2. Markdown Format:} This constraint requires the entire response to follow specific Markdown formats. Examples of this constraint:
\begin{itemize}
    \item The response must include at least/at most/exactly \{N\} headers at level \{M\}.
    \item The response must include at least/at most/exactly \{N\} ordered/unordered lists. Each list must include at least/at most/exactly \{M\} items.
    \item The response must include at least/at most/exactly \{N\} code blocks formatted with triple backticks (\verb|```|) and a specified language (e.g., \verb|```python|).
    \item The response must include at least/at most/exactly {N} horizontal rules, formatted as \texttt{---} or \texttt{***}.
    \item The response must include at least/exactly {N} hyperlinks formatted as \texttt{[text](URL)}.
\end{itemize}

\paragraph{3. Bullet Format:} This constraint specifies the requirements for bullet points. Examples of this constraint:
\begin{itemize}
    \item Format specific content (e.g., reasons, contributions, purposes, and names) into a bulleted list containing at least/at most/exactly \{N\} points.
    \item Each bullet point must include at least/at most/exactly \{N\} words/sentences.
    \item Each bullet point must begin/end with a specific keyword/phrase: \{keyword/phrase\}.
\end{itemize}

\paragraph{4. Language Format:} This constraint specifies the language requirements. Examples of this constraint:
\begin{itemize}
    \item The entire response must be written exclusively in \{language, such as Chinese, English\}.
    \item The response must include a \{language, such as Chinese, English\} idiom/ancient poem.
\end{itemize}

\paragraph{5. Case Sensitivity:} This constraint defines the required case for words. Examples of this constraint:
\begin{itemize}
    \item Write all words in lowercase/uppercase case.
    \item The response must include at least/at most/exactly \{N\} lowercase/uppercase words.
\end{itemize}

\subsection{Style Constraints}
\paragraph{1. Rhetorical Style Constraints:} This constraint specifies the rhetorical style to be used. Examples of this constraint:
\begin{itemize}
    \item Include rhetorical questions to engage the audience.
    \item Conclude with a strong call to action.
\end{itemize}

\paragraph{2. Tone and Emotion Constraints:} This constraint specifies the tone or emotion of the response. Examples of this constraint:
\begin{itemize}
    \item Write the response in a \{tone/emotion, e.g., positive/neutral/negative/academic/persuas- ive/humorous/sarcastic\} style suitable for a \{field/topic, e.g., motivational speech\}.
    \item Use short, punchy sentences to create urgency and excitement.
    \item Convey empathy/sincerity/urgency in the response.
    \item Use a neutral/optimistic/pessimistic tone throughout the response.
    \item The response must be academic/persuasive/humorous/sarcastic.
\end{itemize}

\paragraph{3. Voice Constraints:} This constraint specifies whether the response should use active or passive voice. Examples of this constraint:
\begin{itemize}
    \item Write the response in active/passive voice.
    \item The response must include at least/at most/exactly \{N\} sentences in passive/active voice.
\end{itemize}

\paragraph{4. Sentence Structure Constraints:} This constraint specifies the complexity or structure of sentences. Examples of this constraint:
\begin{itemize}
    \item The response/n-th paragraph must include at least/at most/exactly \{N\} simple/compound/complex sentences.
\end{itemize}

\paragraph{5. Sentence Type Constraints:} This constraint specifies the type of sentences. Examples of this constraint:
\begin{itemize}
    \item The response/n-th paragraph must include at least/at most/exactly \{N\} declarative/interrogative/exclamatory/imperative sentences.
\end{itemize}

\paragraph{6. Readability Constraints:} This constraint specifies the readability of the response. Examples of this constraint:
\begin{itemize}
    \item Tailor the response for specific audience (e.g., children, laypersons, professionals, experts).
    \item Use at least/at most/exactly \{N\} technical terms related to \{field/topic\}.
    \item The response must simplify technical jargon, providing explanations for terms.
    \item Avoid using jargon/slang/archaic words.
\end{itemize}

\paragraph{7. Person Constraints:} This constraint specifies the narrative perspective to be used. Examples of this constraint:
\begin{itemize}
    \item The response must be written in the first/second/third person.
    \item Avoid using personal pronouns.
    \item Include at least \{N\} sentences addressing the reader directly.
\end{itemize}

\paragraph{8. Miscellaneous Style Constraints:} Covers specific stylistic choices not covered above. Examples of this constraint:
\begin{itemize}
    \item Mimic the writing style of \{author/speaker\}.
    \item Include at least \{N\} metaphors/similes in the response.
\end{itemize}
% \clearpage
\section{Conflict Types}\label{section.conflict_types}
\paragraph{1. Conflicts between Content Constraints (CC)}
    \begin{itemize}
        \item Definition: Conflicts occur when two content requirements contradict each other.
        \item Example 1: The itinerary must exclude any mention of national parks vs. The itinerary must include national parks.\\
        Explanation: One constraint says not to mention national parks, while the other requires them to be included.
    \end{itemize}

\paragraph{2. Conflicts between Keyword Constraints (KK)}
\begin{itemize}
    \item Definition: Conflicts arise when keyword-related rules are in opposition.
    \item Example 1: Include the keyword ``like" vs. Avoid using the keyword ``like."\\
    Explanation: The instructions directly contradict each other, as one demands the use of the keyword and the other forbids it.
    \item Example 2: The keyword ``resignation" must appear at least three times vs. Do not include the keyword ``resignation."\\
    Explanation: One rule requires the keyword ``resignation" to appear multiple times, while the other explicitly bans it.
    \item Example 3: The keyword ``strategy" must appear before the keyword ``quality" vs. The keyword ``quality" must appear before the keyword ``strategy."\\
    Explanation: This is a conflict of word order, where one rule demands ``strategy" precedes ``quality" and the other dictates the opposite.
    \item Example 4: Use the keyword ``bank" as a verb vs. Use the keyword ``bank" as a noun.\\
    Explanation: The word ``bank" is given different roles in each constraint, making them incompatible.
    \item Example 5: The keywords ``transaction" and ``clarification" must appear in the same paragraph, with no more than five words separating them vs. The keywords ``transaction" and ``clarification" must appear in the same paragraph, with at least six words separating them.\\
    Explanation: The two rules conflict in terms of the allowable distance between the two keywords.
\end{itemize}

\paragraph{3. Conflicts between Phrase Constraints (PP)}
\begin{itemize}
    \item Definition: Conflicts where different rules dictate how phrases should appear.
    \item Example 1: The first paragraph starts with the phrase ``Embark on an unforgettable journey" vs. The first paragraph starts with the phrase ``Begin an unforgettable journey."\\
    Explanation: The rules conflict in terms of how the first paragraph should begin, requiring different phrases.
    \item Example 2: The first paragraph starts with the phrase ``Embark on an unforgettable journey" vs. Do not include the phrase ``Embark on an unforgettable journey."\\
    Explanation: One rule mandates the phrase to appear at the beginning, while the other forbids its use.
    \item Example 3: The first paragraph starts with the phrase ``Embark on an unforgettable journey" vs. The first paragraph should not start with the phrase ``Embark on."\\
    Explanation: The first rule dictates that the paragraph must start with ``Embark on an unforgettable journey," while the second rule prohibits starting the paragraph with any phrase that begins with ``Embark on." 
\end{itemize}

\paragraph{4. Conflicts between Length Constraints (LL)}
\begin{itemize}
    \item Definition: Conflicts occur when constraints are in opposition regarding the length or size of elements (e.g., word count, number of sentences, number of paragraphs).
    \item Example 1: The email must contain exactly five paragraphs, with each paragraph consisting of at least 80 words vs. The email must contain at most 300 words.\\
    Explanation: If there are exactly five paragraphs with a minimum of 80 words each, the total word count exceeds 300, making the two rules incompatible.
    \item Example 2: The email must contain exactly five paragraphs vs. The email must contain at most four paragraphs.\\
    Explanation: The first rule requires five paragraphs, while the second limits it to four.
    \item Example 3: Each paragraph consists of at least 100 words vs. The first paragraph contains four sentences, with each consisting of at most 20 words.\\
    Explanation: If each sentence in the first paragraph has at most 20 words, the total word count will not exceed 80. This directly conflicts with the rule requiring each paragraph to contain at least 100 words.
    \item Example 4: Each paragraph consists of at least 100 words vs. The first paragraph has between 50 and 80 words.\\
    Explanation: One rule requires the paragraph to have at least 100 words, while the other limits it to a smaller word count.
    \item Example 5: The email must contain exactly five paragraphs, with each paragraph consisting of at least five sentences vs. The email must contain at most 20 sentences.\\
    The first rule requires at least 25 sentences (5 paragraphs × 5 sentences), which conflicts with the second rule, which limits the total to 20 sentences.
\end{itemize}

\paragraph{5. Conflicts between Format Constraints (FF)}
\begin{itemize}
    \item Definition: Conflicts between different formatting requirements.
    \item Example 1: The response must include at least two level-1 headers vs. The response can only use level-2 headers.\\
    Explanation: One rule requires level-1 headers, while the other forbids them, limiting the response to only level-2 headers.
    \item Example 2: The email must be formatted in JSON with the following keys: ``subject", ``body", and ``signature" vs. The email must be formatted in JSON with two keys.\\
    Explanation: One rule requires three keys, while the other allows only two.
    \item Example 3: Present the fine dining recommendations in a bulleted list of exactly three points vs. Avoid using bullet points.\\
    Explanation: The first rule requires bullet points, while the second forbids them.
    \item Example 4: The email must be written exclusively in English vs. The email must include a Chinese idiom.\\
    Explanation: One rule requires the email to be only in English, while the other mandates the inclusion of a Chinese idiom, presented in Chinese.
    \item Example 5: The email should include at least five uppercase words vs. The email must be written in lowercase.\\
    Explanation: One rule requires uppercase words, while the other specifies that everything must be in lowercase.
\end{itemize}

\paragraph{6. Conflicts between Style Constraints (SS)}
\begin{itemize}
    \item Definition: Conflicts arise when different stylistic rules are at odds with each other.
    \item Example 1: The response must be written in the first person vs. The response must be written in the second person.\\
    Explanation: The two rules conflict because one requires a first-person perspective, while the other demands a second-person perspective.
    \item Example 2: Ensure the email is written in a formal tone vs. Ensure the email is written in an informal tone. \\
    Explanation: One rule demands a formal tone, while the other requires an informal one.
    \item Example 3: Tailor the response for laypersons vs. Tailor the response for experts. \\
    Explanation: The two rules conflict because they require the response to be suitable for different audiences: laypersons and experts.
\end{itemize}

\paragraph{7. Conflicts between Keyword Constraints and Phrase Constraints (KP)}
\begin{itemize}
    \item Definition: Conflicts between specific keywords and larger phrases.
    \item Example 1: Refrain from using the keyword ``unforgettable" vs. The first paragraph starts with the phrase ``Embark on an unforgettable journey."\\
    Explanation: The first rule forbids the use of ``unforgettable", while the second requires it as part of a phrase.
    \item Example 2: The first paragraph starts with the phrase ``Embark on an unforgettable journey" vs. The keyword ``unforgettable" must appear before ``embark."\\
    Explanation: The first rule specifies that the phrase ``Embark on an unforgettable journey" should be used, with ``embark" coming first. However, the second rule requires that the word ``unforgettable" must precede ``embark," creating an ordering conflict. 
\end{itemize}

\paragraph{8. Conflicts between Phrase Constraints and Content Constraints (PC)}
\begin{itemize}
    \item Definition: Conflicts arise when specific phrase requirements contradict broader content or thematic requirements, making it impossible to adhere to both at the same time.
    \item Example 1: The email must include the phrase ``Thank you for your business" vs. The email should not express gratitude or appreciation\\
    Explanation: The first rule requires a specific phrase expressing gratitude, while the second rule prohibits any expression of gratitude, making these content and phrase constraints incompatible.
    \item Example 2: The response must contain the phrase ``A visit to Yellowstone National Park is a must" vs. The itinerary must exclude any mention of national parks \\
    Explanation: The first rule mandates mentioning Yellowstone National Park, while the second rule explicitly forbids mentioning any national parks, creating a direct conflict.
    \item Example 3: The introduction must include the phrase ``We guarantee the lowest prices" vs. The content must not make any guarantees or promises\\
    Explanation: The first rule demands a specific phrase that guarantees low prices, while the second rule forbids making guarantees, creating a contradiction.
\end{itemize}

\paragraph{9. Conflicts between Phrase Constraints and Style Constraints (PS)}
\begin{itemize}

    \item Definition: Conflicts arise when specific phrase requirements contradict stylistic requirements, such as tone, perspective, or formality.
    \item Example 1: The response must include the phrase: ``I strongly believe this is the best approach." vs. The response must be written in the second person\\
    Explanation: The phrase uses first-person perspective (``I strongly believe"), but the style constraint requires the response to be in the second person.
    \item Example 2: The response must include the phrase: ``Hey buddy, this is gonna be awesome!" vs. The response must be written in a formal tone\\
    Explanation: The required phrase is informal and conversational, but the style constraint demands a formal tone.
    \item Example 3: The response must include the phrase: ``The stochastic process adheres to a Markovian property." vs. The response must be tailored for laypersons\\
    Explanation: The phrase contains technical jargon suited for experts, but the style constraint requires the response to be accessible to laypersons.
    \item Example 4: The response must include the phrase: ``This is the worst decision ever made." vs. The response must maintain a neutral and unbiased tone\\
    Explanation: The required phrase expresses a strong negative opinion, contradicting the neutrality constraint.
\end{itemize}

\newpage
\section{Prompt Templates}

\begin{table}[h]
\footnotesize
\centering
   \begin{tabular}{
    m{0.48\textwidth}
    }
    \toprule
    I currently have a simple seed instruction (i.e., [Seed Instruction]). Your task is to make it more complex by adding additional constraints.
    To assist you in completing this task, I will provide six types of constraints for your reference (i.e., [Reference Constraints]), which include `Content Constraints', `Keyword Constraints', `Phrase Constraints', `Length Constraints', `Format Constraints', and `Style Constraints'.
    Each type of constraint includes several different sub-constraints. For example, 'Format Constraints' consist of five sub-constraints: JSON Format, Markdown Format, Bullet Format, Language Format, and Case Sensitivity. 
    For each sub-constraint, we will first provide a definition followed by example templates. You may choose a suitable template from these examples or create your own, as long as it satisfies the sub-constraint's definition. Ensure that you strictly apply all sub-constraints from each type without prioritizing any particular one.\\
    \\
    Below is [Reference Constraints].\\
    \textbf{[Reference Constraints]}\\
    \{\textit{Constraints in \S \ref{section.constraint_dimensions}}\}.\\
    \\
    Below is the requirements for modifying the seed instruction.\\
    \textbf{[Requirements]:}\\
    \begin{enumerate}
        \item You must use all constraints from [Reference Constraints].
        \item Feel free to use any constraints other than [Reference Constraints] that you deem appropriate.
        \item When adding constraints to the seed instruction, you are free to combine and paraphrase the selected constraints as needed. Seamlessly integrate these constraints into the seed instruction without omitting any key information, and avoid directly listing the selected constraints.
        \item If the seed instruction is a question, please do not modify the seed instruction. Add constraints after the seed instruction.
        \item Directly output the modified instruction (the instruction with added constraints in plain text format, i.e., [Modified Instruction]), without any analysis. The modified instruction must not contain line breaks.
    \end{enumerate}
    \\    
    Below is the seed instruction.\\
    \textbf{[Seed Instruction]:} \{\textit{Seed Instruction}\}\\
    \textbf{[Modified Instruction]:} \\
    \bottomrule
    \end{tabular}
\caption{Prompt template for expanding seed instructions.}
\label{table.prompt.expand_instruction}
\end{table}

\begin{table}
\footnotesize
\centering
   \begin{tabular}{
    m{0.48\textwidth}
    }
    \toprule
    I currently have an instruction (i.e., [Instruction]) that includes multiple constraints, all of which can be satisfied simultaneously.
    Your task is to add new constraints to this instruction. These new constraints should conflict with the existing ones in the given instruction, meaning they cannot be satisfied at the same time. However, the new constraints themselves must not conflict with one another.
    To assist you in completing this task, I will provide six types of constraints for your reference (i.e., [Reference Constraints]), which include `Content Constraints', `Keyword Constraints', `Phrase Constraints', `Length Constraints', `Format Constraints', and `Style Constraints'.
    Each type of constraint includes several different sub-constraints. For example, `Format Constraints' consist of five sub-constraints: JSON Format, Markdown Format, Bullet Format, Language Format, and Case Sensitivity.\\
    \\
    Below is [Reference Constraints].\\
    \textbf{[Reference Constraints]}\\
    \{\textit{Constraints in \S \ref{section.constraint_dimensions}}\}.\\
    \\
    Below are the types of conflicts and their corresponding examples:\\
    \textbf{[Conflict Types]}\\
    \{\textit{Conflict types in \S \ref{section.conflict_types}}\}.\\
    \\
    When adding conflicting constraints into the given instruction, you must adhere to the following requirements:\\
    \textbf{[Requirements]:}
    \begin{enumerate}
        \item Select constraints exclusively from [Reference Constraints].
        \item The selected constraints must conflict with at least one constraint in the provided instruction. However, there must be no internal conflicts among the selected constraints. Do not include constraints that do not conflict with the given instruction’s existing constraints.
        \item The conflicts between the selected constraints and the existing constraints in the given instruction must be explicit and unambiguous.
        \item The conflicts must cover all nine conflict types in [Conflict Types], with each conflict type containing one pair of conflicting constraints.
        \item The conflicting examples in [Conflict Types] and the example below are provided to aid your understanding. However, I would prefer that you not rely solely on these examples. Please come up with a wider variety of conflict scenarios.
        \item The output must strictly adhere to the format shown in the example below.
    \end{enumerate}
    
    Here is an example:\\
    \{\textit{In-Context Example}\}\\
    \\
    \textbf{[Instruction]:} \{\textit{Instruction}\}\\
    \bottomrule
    \end{tabular}
\caption{Prompt template for adding conflicting instructions.}
\label{table.prompt.add_conflict}
\end{table}

\begin{table}
\footnotesize
\centering
   \begin{tabular}{
    m{0.48\textwidth}
    }
    \toprule
    I currently have an instruction (i.e., [Instruction]) that includes multiple constraints. Please help me judge whether the instruction contains conflicting constraints, which cannot be satisfied simultaneously.\\
    \\
    Answer ``YES" if you think the [Instruction] contains conflicting constraints or requirements; otherwise, answer ``NO."\\
    \\
    Please first provide an explanation and then give the answer. Your output should follow the format below:\\
    \textbf{[Explanation]}: Your explanation\\
    \textbf{[Answer]}: YES / NO\\
    \\
    \textbf{[Instruction]:} \{\textit{Instruction}\}\\
    \bottomrule
    \end{tabular}
\caption{Prompt template for conflict detection.}
\label{table.prompt.conflict_detection}
\end{table}

\begin{table}[]
\footnotesize
\centering
   \begin{tabular}{
    m{0.48\textwidth}
    }
    \toprule
    You will be given an Instruction containing multiple constraints, some of which are conflicting and cannot be simultaneously satisfied. You will also be provided with a Model Response, which is generated based on the given Instruction.\\
    \\
    Your task is to evaluate the Model Response and determine whether the model has recognized the conflicting constraints within the Instruction when generating its response. Follow these evaluation rules:
    \begin{enumerate}
    \item If the model does not acknowledge the conflicting constraints and directly provides a response to the instruction, output 1.
    % \item If the model identifies the conflicting constraints and, as a result, does not provide a response to the instruction, but it also does not ask the user for clarification, output 2.
    \item If the model identifies the conflicting constraints, does not provide a response to the instruction, and explicitly asks the user to clarify the instruction, output 2.
    \item If the model identifies the conflicting constraints, resolves them on its own, and then provides a response to the resolved instruction, output 3.
    \item If the model’s response does not fit into any of the above categories, output 4.
    \end{enumerate}
    Your output must strictly follow this format:\\
    \textbf{\#\# Output Format} \\
    **Analysis**: A brief analysis of the Model Response.  \\
    **Answer**: The evaluation result (1, 2, 3, or 4) without any additional content.\\
    \\
    \textbf{\#\# Evaluation Input}\\
    **Model Response**\\
    \{\textit{Model Response}\}\\
    \\
    **Instruction**\\
    \{\textit{Instruction}\}\\
    \\
    \textbf{\#\# Evaluation Output}\\
    \\
    \bottomrule
    \end{tabular}
\caption{Prompt template for evaluating LLM behavior when handling conflicting instructions. }
\label{table.prompt.conflict_resolution_behavior}
\end{table}

\begin{table}
\footnotesize
\centering
   \begin{tabular}{
    m{0.48\textwidth}
    }
    \toprule
    I currently have an instruction (i.e., [Instruction]) that includes multiple constraints. Please analyze whether any of these constraints are in conflict and cannot be satisfied simultaneously. If no such conflicts are found, proceed to execute the instruction. However, if conflicting constraints are detected, refrain from executing the instruction. Instead, clearly indicate the presence of conflicting constraints and request further clarification.\\
    \\
    \textbf{[Instruction]:} \{\textit{Instruction}\}\\
    \bottomrule
    \end{tabular}
\caption{Prompt template for instruction-following with an expected conflict resolution behavior.}
\label{table.prompt.instruction_following_behavior}
\end{table}

\begin{table}
\footnotesize
\centering
   \begin{tabular}{
    m{0.48\textwidth}
    }
    \toprule
    Your task is to act as a quality evaluator, analyzing the content of the Model Response to determine whether it fully satisfies the requirements outlined in the Instruction.
    When evaluating, you should adhere to the following judgment criteria:\\
    1. Answer ``YES" if the Model Response entirely fulfills all the requirements specified in the instruction.\\
    2. Answer ``NO" if the Model Response fails to meet all the requirements or provides no relevant information for the given instruction.\\
    \\
    \#\# \textbf{Evaluation Steps}\\
    Please analyze the Model Response and Instruction carefully, adhering to the following steps:\\
    Step 1: Analyze the Instruction, then extract relevant content from the Model Response. Copy sentences from the Model Response exactly as they are, without any modification.
    \begin{enumerate}
        \item If the instructions include constraints related to keywords or phrases (e.g., keyword definitions, keyword frequency, or phrase frequency), extract the sentences containing the specified keywords or phrases. Record the positions of these keywords or phrases, if necessary.
        \item If the instructions include constraints related to specific information or topics, extract segments containing the relevant information or topics.
        \item If the instructions include constraints related to output formats or styles, extract segments that reflect the specified formats or styles.
    \end{enumerate}
    Step 2: Analyze whether the Instruction’s constraints are fully satisfied.\\
    Step 3: Provide your evaluation answer (``YES" or ``NO"), without adding extra content.\\
    \\
    \textbf{\#\# Output Format}\\
    **Step 1**: The extracted content from the Model Response.\\
    **Step 2**: A brief analysis of the Instruction.\\
    **Step 3**: YES / NO\\
    \\
    \textbf{\#\# Evaluation Input}\\
    **Model Response**\\
    \{\textit{Model Response}\}\\
    \\
    **Instruction**\\
    \{\textit{Instruction}\}\\
    \\
    \textbf{\#\# Evaluation Output}
    \\
    \bottomrule
    \end{tabular}
\caption{Prompt template for instruction-following evaluation.}
\label{table.prompt.instruction_following}
\end{table}

\begin{table}
\footnotesize
\centering
   \begin{tabular}{
    m{0.48\textwidth}
    }
    \toprule
    You will be provided with the following:
    \begin{enumerate}
        \item Two instructions, Instruction 1 and Instruction 2.
        \item A Model Response, generated by a model using Instruction 1 and Instruction 2.
    \end{enumerate}
    
    Your task is to act as a quality evaluator, analyzing the content of the Model Response to determine which instruction’s all constraints is fully satisfied based on the following rules:
    \begin{enumerate}
        \item If all constraints in Instruction 1 are fully satisfied, output 1.
        \item If all constraints in Instruction 2 are fully satisfied, output 2.
        \item If neither instruction’s constraints are fully satisfied, output -1.
    \end{enumerate}
    
    Note that these two instructions contain conflicting constraints, making it impossible for the Model Response to fully satisfy both simultaneously.\\
    \\
    \textbf{\#\# Evaluation Steps}\\
    Please analyze the Model Response, Instruction 1 and Instruction 2 carefully, adhering to the following steps:\\
    Step 1: Analyze Instruction 1 and Instruction 2, and then extract relevant content from the Model Response. You should copy sentences from the Model Response exactly as they are, without any modification.
    \begin{enumerate}
        \item If the instructions include constraints related to keywords or phrases (e.g., keyword definitions, keyword frequency, or phrase frequency), extract the sentences containing the specified keywords or phrases. Record the positions of these keywords or phrases, if necessary.
        \item If the instructions include constraints related to specific information or topics, extract segments containing the relevant information or topics.
        \item If the instructions include constraints related to output formats or styles, extract segments that reflect the specified formats or styles.
    \end{enumerate}
    Step 2: Analyze which instruction’s constraints are fully satisfied.\\
    Step 3: Directly give your evaluation answer (1, 2, or -1) without any additional content.\\
    \\
    \textbf{\#\# Output Format}\\
    **Step 1**: The extracted content from the Model Response.\\
    **Step 2**: A brief analysis of Instruction 1 and Instruction 2.\\
    **Step 3**: The evaluation result (1, 2, or -1) without any additional content.\\
    \\
    \textbf{\#\# Evaluation Input}\\
    **Model Response**\\
    \{\textit{Model Response}\}\\
    \\
    **Instruction 1**\\
    \{\textit{Instruction 1}\}\\
    \\
    **Instruction 2**\\
    \{\textit{Instruction 2}\}\\
    \\
    \textbf{\#\# Evaluation Output}\\
    \\
    \bottomrule
    \end{tabular}
\caption{Prompt template used to evaluate which of the two mutually conflicting instructions is satisfied by a model's output. }
\label{table.prompt.conflict_resolution}
\end{table}

\newpage
\section{LLM Response Evaluation}

\subsection{Instruction-Following Evaluation}
Table \ref{table.consistency_instruction_following} presents the consistency between GPT-4o and human evaluations across different types of instruction-following constraints. The results indicate that GPT-4o's automatic evaluation aligns closely with human judgment, confirming the reliability of the automated assessment.

\begin{table}[h]
    \scriptsize
    \centering
    \setlength{\tabcolsep}{3.5pt}% column space
    % \resizebox{0.48\textwidth}{!}{
    \begin{tabular}{l|cccccc}
    % \toprule
    \hline
    \textbf{Constraints }&\textbf{Content}&\textbf{Keyword}&\textbf{Phrase}&\textbf{Style}&\textbf{Length}&\textbf{Format}\\
    % \midrule
    \hline
    Consistency & 92\% & 88\% & 100\% & 92\% & 96\% & 100\%\\
    % \bottomrule
    \hline
    \end{tabular}%
    % }
    % \vspace{-2mm}
    \caption{
    Consistency between GPT-4o and human evaluations across different types of constraints in instruction-following.
    }
    % \vspace{-4mm}
    \label{table.consistency_instruction_following}
\end{table}

\subsection{Conflict Resolution Evaluation}
In Figure \ref{fig.conflict_resolution_density}, we employ GPT-4o to analyze the conflict resolution behaviors of LLMs. We classify LLMs' responses into four types and employ GPT-4o to assign behavior labels to LLMs' responses. 
To ensure annotation quality, we also utilize Gemini-2.5-Pro and perform manual labeling on a subset of the LLM responses. 
Due to budget constraints, the evaluation was conducted on a sample of 400 responses sourced from four LLMs. Specifically, we randomly selected 100 responses from each model, covering cases where the instructions contained $k$ conflicts, with $k$ ranging from 1 to 6.

As shown in Table \ref{table.consistency_conflict_resolution}, both GPT-4o and Gemini-2.5-Pro exhibit strong agreement with human annotations, with GPT-4o achieving 98\% accuracy. Furthermore, the inter-annotator agreement between GPT-4o and Gemini-2.5-Pro, measured by Cohen's Kappa is 0.746—indicating substantial consistency according to Landis \& Koch \cite{Landis1977TheMO}.

We further analyze Gemini-2.5-Pro's annotation results in Table \ref{table.gemini}, which align with GPT-4o's findings in Table \ref{table.gpt4}:\\
1. GPT-4o and Qwen2.5-32B predominantly exhibit Behavior 1, failing to acknowledge conflicts in their responses. \\
2. Claude-3.5 models partially address conflicts, with Claude-3.5-Sonnet showing the highest tendency to acknowledge conflicts (Behavior 2 and Behavior 3).

\begin{table}[h]
    \scriptsize
    \centering

    \begin{tabular}{l|c|c}
    \hline
    \textbf{LLMs' Responses / Annotators} & \textbf{GPT-4o} & \textbf{Gemini-2.5-Pro} \\
    \hline
    Claude-3.5-Sonnet & 98 & 97 \\
    Claude-3.5-Haiku & 94 & 82 \\
    GPT-4o & 100 & 90 \\
    Qwen2.5-32B & 100 & 89 \\
    \hline
    \textbf{Average} & \textbf{98.0} & \textbf{89.5} \\
    \hline
    \end{tabular}
    \caption{Consistency (\%) between LLM annotators and human evaluations in conflict resolution behavior analysis.}
    \label{table.consistency_conflict_resolution}
\end{table}

\begin{table}[t]
    \centering
    \footnotesize
    \begin{tabular}{l|c|c|c|c}
    \hline
    \textbf{LLMs' responses / Behaviors} & \textbf{B1} & \textbf{B2} & \textbf{B3} & \textbf{B4} \\
    \hline
    Claude-3.5-Sonnet & 28 & 30 & 41 & 1 \\
    \hline
    Claude-3.5-Haiku & 72 & 11 & 13 & 4 \\
    \hline
    GPT-4o & 96 & 4 & 0 & 0 \\
    \hline
    Qwen2.5-32B & 100 & 0 & 0 & 0 \\
    \hline
    \end{tabular}
    \caption{The distribution of conflict resolution behaviors across different LLMs, as annotated by GPT-4o (400 responses).}
    \label{table.gpt4}
\end{table}

\begin{table}[t]
    \centering
    \footnotesize
    \begin{tabular}{l|c|c|c|c}
    \hline
    \textbf{LLMs' responses / Behaviors} & \textbf{B1} & \textbf{B2} & \textbf{B3} & \textbf{B4} \\
    \hline
    Claude-3.5-Sonnet & 24 & 30 & 46 & 0 \\
    \hline
    Claude-3.5-Haiku & 54 & 9 & 37 & 0 \\
    \hline
    GPT-4o & 86 & 4 & 10 & 0 \\
    \hline
    Qwen2.5-32B & 89 & 0 & 11 & 0 \\
    \hline
    \end{tabular}
    \caption{The distribution of conflict resolution behaviors across different LLMs, as annotated by Gemini-2.5-Pro (400 responses).}
    \label{table.gemini}
\end{table}

\end{document}